\documentclass[sigplan,screen,nonacm]{acmart}

\renewcommand\footnotetextcopyrightpermission[1]{}

\usepackage[colorinlistoftodos]{todonotes}
\usepackage{romannum} 
\usepackage[flushleft]{threeparttable}
\usepackage{multirow}
\usepackage{booktabs}
\usepackage{array}
\usepackage{subcaption}
\usepackage{comment}
\usepackage{appendix}

\usepackage{float}
\usepackage[linesnumbered, ruled]{algorithm2e}

\usepackage{caption}
\usepackage{amsmath}
\usepackage{xcolor,cancel}

\usepackage{xcolor}
\usepackage{listings}
\definecolor{xschedkeyword}{HTML}{008080}
\definecolor{xschedcomment}{HTML}{B22222}

\lstdefinestyle{xschedstyle}{
    language=Python,
    basicstyle=\ttfamily\small,
    keywordstyle=\color{xschedkeyword}\bfseries,
    commentstyle=\color{xschedcomment},
    numberstyle=\scriptsize\color{black},
    numbers=left,
    numbersep=8pt,
    breaklines=true,
    frame=none,
    showstringspaces=false,
    morekeywords={void, then, TRUE, FALSE, in}
}

\AtBeginDocument{%
  \providecommand\BibTeX{{%
    \normalfont B\kern-0.5em{\scshape i\kern-0.25em b}\kern-0.8em\TeX}}}

\begin{document}

\title{Strait: Perceiving Priority and Interference in ML Inference Serving}

 


\author{Haidong Zhao}
\orcid{0009-0009-9864-5520}
\affiliation{
  \institution{Inria \& Sorbonne University}
  \city{Paris}
  \country{France}
}

\author{Nikolaos Georgantas}
\orcid{0000-0001-5704-4889}
\affiliation{
  \institution{Inria}
  \city{Paris}
  \country{France}
}


\begin{abstract}
Machine learning (ML) inference serving systems host deep neural network (DNN) models and schedule incoming inference requests across deployed GPUs.
However, limited support for task prioritization and insufficient latency estimation under concurrent execution may restrict their applicability in on-premises scenarios.
We present \emph{Strait}, a serving system designed to enhance deadline satisfaction for dual-priority inference traffic under high GPU utilization.
To improve latency estimation, Strait models potential contention during data transfer and accounts for kernel execution interference through an adaptive prediction model.
By drawing on these predictions, it performs priority-aware scheduling to deliver differentiated handling.
Evaluation results under intense workloads suggest that Strait reduces deadline violations for high-priority tasks by 1.02 to 11.18 percentage points while incurring acceptable costs on low-priority tasks. 
Compared to software-defined preemption approaches, Strait also exhibits more equitable performance.
\end{abstract}

\maketitle

\section{Introduction}
\label{sec:strait_introduction}

Machine learning (ML) techniques and \emph{already-trained} deep neural network (DNN) models are deployed to power intelligent applications. 
Since ML inferences are both resource-intensive and latency-sensitive workloads, accelerators such as GPUs\footnote{This paper adopts terminology from NVIDIA GPUs}, which offer massive parallel processing capabilities~\cite{dally_evolution_2021}, are employed to meet these demands. 
ML inference serving systems can multiplex system resources across deployed DNN models to enhance cost efficiency while delivering satisfactory performance.
Nonetheless, existing systems may exhibit suboptimal performance across many on-premises scenarios.


First, scheduling inference requests by \emph{priority levels} may be insufficiently supported, as existing inference serving systems are primarily tailored to online services.
Consequently, these systems typically aim to optimize overall performance in terms of target metrics~\cite{gujarati_serving_2020,wu_irina_2020,hu_scrooge_2021,shen_nexus_2019,crankshaw_inferline_2020,choi_serving_2022,crankshaw_clipper_2017}.
However, certain scenarios can exhibit varying priority levels~\cite{han_microsecond-scale_2022,noghabi_emerging_2020,shen_xsched_2025}, urgency, and sensitivity to deadline violations.
For example, in industrial monitoring, tasks such as \emph{quality control}~\cite{xu_high-throughput_2023,audis_edge,huang_automated_2015,vaish_case_2023,jin_glass_2023}—inspecting components for defects—and \emph{equipment health monitoring}~\cite{bose_adepos_2019,cline_predictive_2017,sipos_log-based_2014,davari_predictive_2021} are deemed more critical than workloads such as indoor temperature control.
Additionally, detecting unsafe operations or intrusions in workplaces or production premises may take precedence over other workloads~\cite{zhang_design_2015, tran_human_2023}.

Second, the growing adoption of \emph{on-premises} servers~\cite{vasisht_farmbeats_2017, padmanabhan_gemel_2023, noghabi_emerging_2020} for timely, reliable processing~\cite{adarsh_coverage_2021, xu_high-throughput_2023} makes resource optimization more critical than in cloud environments~\cite{millnert_holoscale_2020, bhardwaj_cilantro_2023, rzadca_autopilot_2020}.
Serving systems typically employ \emph{batching} to consolidate multiple requests for the same model, thereby improving GPU utilization by amortizing off-chip memory accesses.
\emph{Temporal sharing}~\cite{gujarati_serving_2020, crankshaw_clipper_2017, shen_nexus_2019, crankshaw_inferline_2020} processes batches sequentially, providing relatively predictable inference latency. However, it may be susceptible to Head-of-Line (HOL) blocking, which can lead to GPU underutilization.
Conversely, \emph{spatial sharing} leverages concurrency to enhance GPU utilization~\cite{yu_survey_2022, choi_serving_2022, dhakal_gslice_2020, gilman_characterizing_2021, wesolowski_datacenter-scale_2021, romero_infaas_2021}; however, it introduces interference during concurrent kernel execution and data transfers, impeding latency estimation and potentially compromising deadline satisfaction.
The economic impact of deadline violations has been quantitatively studied in online services; e.g., Amazon reports that every 100 ms of additional latency results in a 1\% loss in sales~\cite{latency_amazon,latency_deloitte}. 
In on-premises scenarios, the consequences can also be severe.
For instance, in high-throughput manufacturing, defective items may escape certain quality inspections, as interruptions in the manufacturing process would incur noticeable production losses, e.g., \$22,000 per minute in the automotive sector~\cite{downtime_cost_1,latency_visual_inspection_1}. 
However, allowing defective items to reach customers may ultimately undermine both economic outcomes and brand reputation.
Given that a GPU can serve hundreds to thousands of requests per second, even a 1 percentage-point (pp) drop in deadline satisfaction can cause deadline-violated requests to scale with the number of GPUs and the duration of serving.

Nevertheless, existing techniques for supporting task prioritization under high GPU utilization are limited at both the hardware and software levels.
Regarding internal GPU scheduling, CUDA stream priority~\cite{CUDA_C++_Programming_Guide} provides a hardware-level hint to favor operations in higher-priority streams. However, its efficacy as a standalone solution remains limited; concurrent kernels can still contend for shared GPU resources, and depending on workload characteristics, high-priority (HP) tasks may experience even more severe interference than low-priority (LP) tasks (Section~\ref{sec:strait_motivation_priority}).
Regarding task scheduling, priority scheduling can prioritize HP tasks~\cite{tf_serving_shared_schedulier, zhang_shepherd_2023, triton_inference_server}; however, its efficacy may be undermined by interference from currently executing or subsequently scheduled tasks, which can still cause HP tasks to miss their deadlines.
Consequently, a critical concern remains: whether scheduling a new batch would violate the deadlines of ongoing tasks or fail to meet its own deadline.
Worse, commodity GPUs may not support \emph{true} task-level preemption~\cite{strati_orion_2024,han_microsecond-scale_2022,otterness_amd_2020,shen_xsched_2025, ng_paella_2023}, and inference is often executed as a compiled computation graph~\cite{tensorrt_doc, onnx_runtime}.
These factors virtually eliminate any opportunity for remediation once scheduling decisions are dispatched.
Accordingly, effective task prioritization and deadline satisfaction are inherently constrained.

To this end, we introduce \emph{Strait}, an ML inference serving system designed to \emph{enhance} deadline satisfaction for dual-priority inference traffic under high GPU utilization.
Strait accounts for potential interference, task priority levels, and runtime overheads when making scheduling decisions.
First, we model interference in data transfer based on the observation that PCIe data transfer with pinned memory follows a FIFO process.
Subsequently, we propose a prediction model for kernel execution interference.
This model is grounded in GPU characteristics and posits that interference exhibits an exponential growth pattern with increasing resource pressure.
Moreover, this prediction model can be dynamically updated using inference feedback, as static models may suffer from performance degradation when workload characteristics drift or when directly applied to different GPU types.
Finally, we design a priority-aware scheduling algorithm that integrates interference prediction to guide its decisions.
Its objective is to meet the deadline of the scheduled batch while reducing deadline violations for running batches to the extent possible.

To summarize, we make the following contributions:
In Section~\ref{sec:strait_motivation}, we demonstrate that task prioritization in existing serving systems is limited, as are interference prediction approaches for inference workloads.
In Section~\ref{subsec:Strait_interference_prediction}, we model data transfer interference and propose an adaptive prediction model to estimate kernel execution interference.
In Section~\ref{subsec:Strait_priority_aware_scheduling}, we design a priority-aware scheduling method that explicitly manages interference at scheduling time.
In Section~\ref{sec:strait_evaluation}, we present evaluation results showing that Strait reduces deadline violations for HP tasks by 1.02 to 11.18 pp, while incurring an acceptable performance trade-off for LP tasks.
We further evaluate the accuracy, limitations, and adaptability of the interference prediction approach, assessing its sustainability using realistic production traces~\cite{shahrad_serverless_2020}.

\section{Background and Motivation}
\label{sec:strait_motivation}

At runtime, executing a batch is equivalent to executing a sequence of \emph{kernels}, which are functions specifically designed to exploit the GPU’s parallel resources to accelerate DNN operations. 
Spatial sharing executes batches concurrently, further improving utilization relative to the sequential execution of temporal sharing.
This is because temporal sharing is limited by timing constraints when forming larger batches to saturate GPU resources, incurs stalls from sequential operations (e.g., data transfer and kernel execution), and can lead to HOL blocking.
However, spatial sharing introduces inevitable interference, as concurrent batches can contend for shared resources. 
Additionally, developers lack fine-grained control once kernels are dispatched to non-preemptive GPUs~\cite{otterness_amd_2020, strati_orion_2024, han_microsecond-scale_2022, shen_xsched_2025}, and internal GPU scheduling mechanisms remain proprietary~\cite{amert_gpu_2017, gilman_demystifying_2021, yang_avoiding_2018}.
Furthermore, inference workloads are often compiled into computation graphs~\cite{tensorrt_doc, onnx_runtime}, where execution proceeds continuously from the first kernel to the last without interruption.

We next examine the limitations of task prioritization techniques in existing serving systems, as well as interference prediction approaches. 
The experiments are conducted on an NVIDIA L4 GPU with the Ada Lovelace architecture, unless stated otherwise.

\subsection{Scheduling Inference with Priorities}
\label{sec:strait_motivation_priority}

When scheduling inference requests on GPUs, existing inference serving systems may exhibit suboptimal task prioritization.
For instance, TensorFlow Serving (TFS) includes a priority queue mechanism for task scheduling in its codebase~\cite{tf_serving_shared_schedulier}; however, to our knowledge, this feature has not been officially released. 
NVIDIA’s Triton Inference Server~\cite{triton_inference_server} explicitly supports configuring the CUDA stream priority~\cite{CUDA_C++_Programming_Guide} mechanism to influence internal GPU scheduling.
To evaluate these mechanisms, we adopt the TFS design and employ task priority scheduling and CUDA stream priority. 
TFS scans each model’s task queue in a round-robin manner and submits a batch once it meets scheduling criteria ~\cite{tf_serving_shared_schedulier}; we replace this with First-Come, First-Served (FCFS), as our empirical results indicate that it reduces tail latency and improves deadline satisfaction.
We co-locate an HP task with an LP task in each scenario and restrict the maximum batch concurrency to two, allowing co-location only between batches from different models.

\begin{figure}[t]
  \centering
  \begin{subfigure}[t]{0.49\linewidth}
    \centering
    \includegraphics[width=\linewidth]{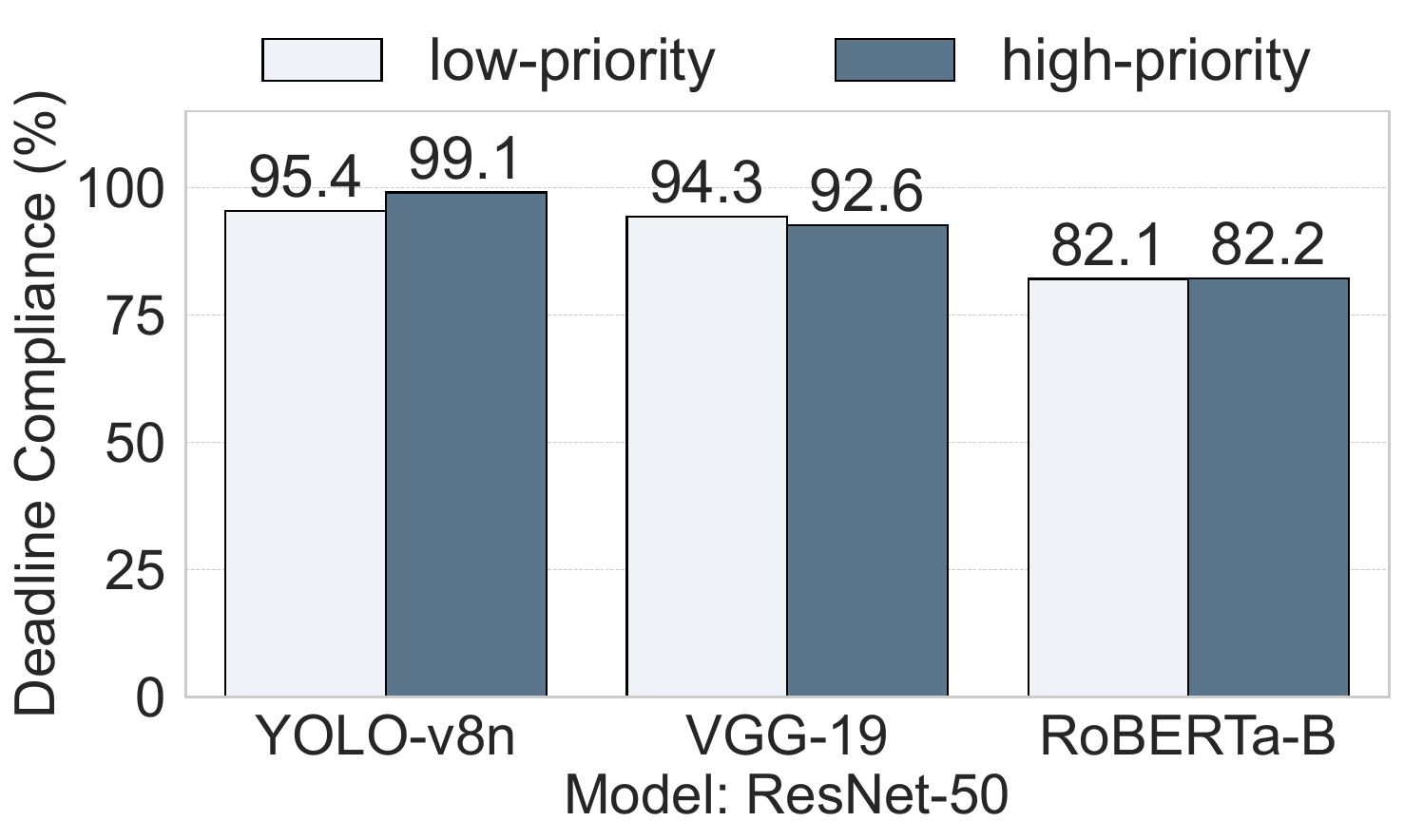}
    \caption{Deadline satisfaction}
    \label{figure:Strait_motivation_deadline_a}
  \end{subfigure}
  \hfill
  \begin{subfigure}[t]{0.49\linewidth}
    \centering
    \includegraphics[width=\linewidth]{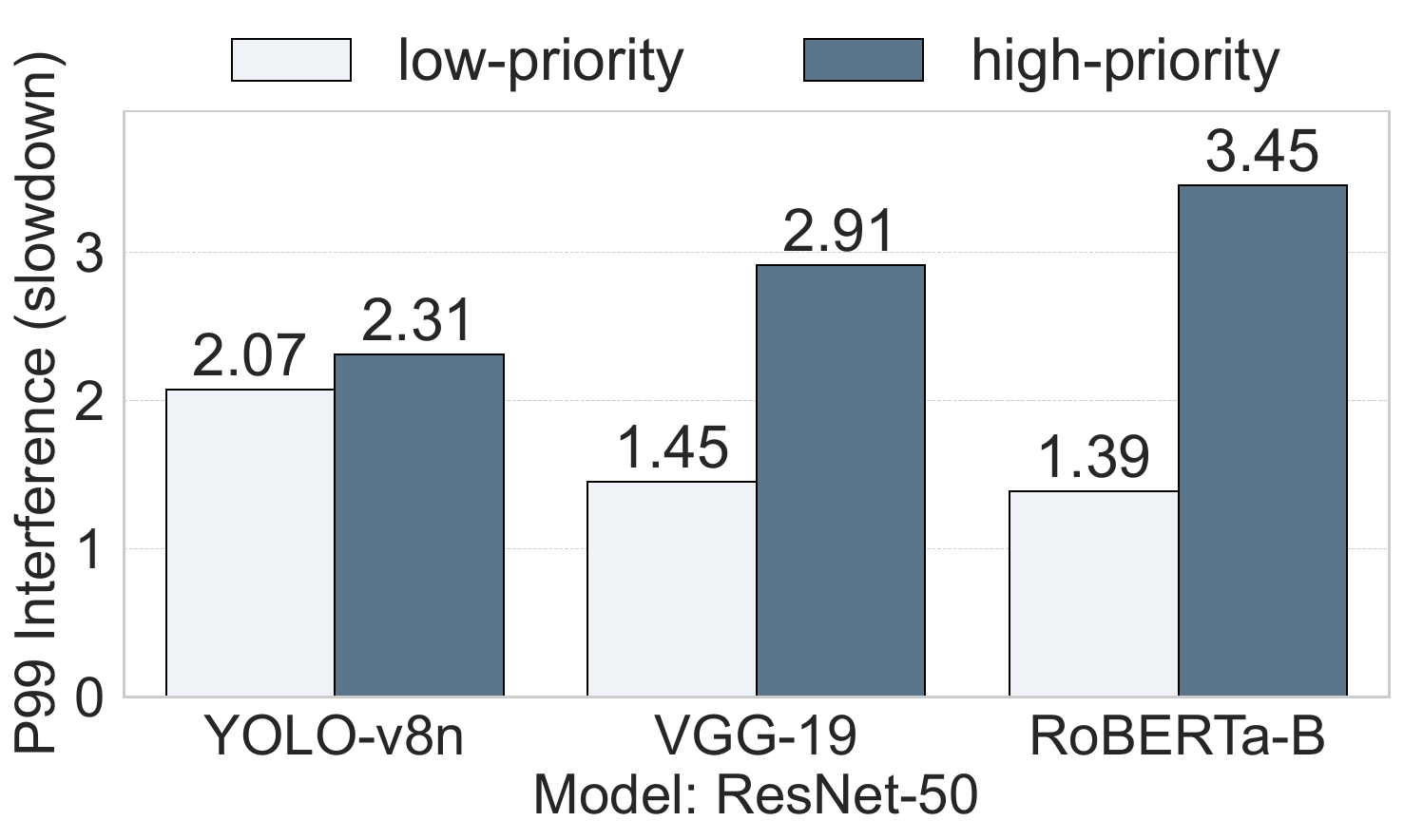}
    \caption{Kernel execution interference}
    \label{figure:Strait_motivation_deadline_b}
  \end{subfigure}
  \caption{
    In each scenario, ResNet-50 (HP task) is co-located with a distinct model (LP task).}
  \label{figure:Strait_motivation_deadline}
\end{figure}

Figure~\ref{figure:Strait_motivation_deadline} presents the results, where a ResNet-50~\cite{he_deep_2016} is sequentially co-located with a relatively lightweight YOLO-v8n~\cite{jocher_ultralytics_2023}, a compute-intensive VGG-19~\cite{simonyan_very_2015}, and a memory-intensive RoBERTa-B~\cite{liu_roberta_2019}.
Across the scenarios, the request arrival rates follow a 3:2:1 ratio, with the load evenly balanced between the co-located models at each priority level.
Although the request volume for ResNet-50 decreases across the scenarios, its deadline compliance rate progressively declines (Figure~\ref{figure:Strait_motivation_deadline_a}).
Additionally, despite using prioritization techniques to favor ResNet-50, its \emph{tail} kernel execution interference (quantified as the slowdown relative to its profiled p95 isolated kernel execution latency, normalized to 1) can exceed that of the co-located model. 
This effect is especially pronounced when paired with heavier models (Figure~\ref{figure:Strait_motivation_deadline_b}).
We utilize Nsight Systems~\cite{nvidia_nsight_system} to profile the interactions among concurrent kernels from distinct batches.
We investigate a case in which a batch-1 (batch size of 1) ResNet-50 experiences severe interference when co-located with a batch-4 RoBERTa-B.
Compared with isolated execution, the total kernel execution latency slowdown exceeds $3.4\times$, the total inter-kernel intervals exceed $6.6\times$, and consequently the overall slowdown 
exceeds $3.6\times$.
In contrast, when this batch-1 ResNet-50 is co-located with a batch-2 YOLO-v8n, these values drop to $1.7\times$, $2.73\times$, and $1.79\times$, respectively.
This discrepancy arises because the resource demands of co-located kernels from RoBERTa-B are considerably higher than those from YOLO-v8n, requiring $2.2\times$ more avg. registers per thread, $2.7\times$ more avg. shared memory, and $3.5\times$ more total threads, among other resources.
Regarding the increase in kernel execution latency, co-location with a resource-intensive kernel or one with similar resource demands can lead to severe interference~\cite{strati_orion_2024}.
Additionally, increasing intervals arise between consecutive kernels, owing to the fact that a kernel's thread blocks must be provisioned with sufficient resources—such as registers and shared memory—before they can be scheduled~\cite{amert_gpu_2017, gilman_demystifying_2021, yang_avoiding_2018, wu_flep_2017}.
Although larger batch sizes improve GPU utilization, they naturally incur longer execution times, thereby severely constraining the time budget available to tolerate potential severe and variable interference. Consequently, managing these interference effects becomes critical to improve task prioritization and deadline satisfaction under high GPU utilization.

\subsection{Estimating Interference Effects}
\label{sec:strait_motivation_prediction}

As shown above, interference-induced kernel execution slowdown can exceed $3.45\times$ (Figure~\ref{figure:Strait_motivation_deadline_b}). Notably, these experiments were conducted with the maximum batch concurrency limited to 2.
In production environments, deployed models can be more heterogeneous, and supported concurrency levels could be higher.
Our empirical results indicate that interference can further intensify in such scenarios, and the corresponding distribution exhibits a \emph{long tail}.

Directly modeling kernel execution~\cite{chen_prophet_2017,chen_baymax_2016} to estimate interference effects is complicated, as executing a batch corresponds to running a sequence of kernels with varying resource demands.
This approach would lead to several issues.
First, kernels with complementary resource demands that co-locate may exhibit different interference effects compared to those with mutual resource demands.
For instance, two memory-bound kernels may experience a greater degree of performance degradation than the case where a compute-bound kernel co-locates with a memory-bound one~\cite{strati_orion_2024}.
Moreover, when using a runtime library~\cite{tensorrt_doc,onnx_runtime} to perform inference, the kernels within a batch execute without interruption, making it difficult to determine which kernels are likely to be co-located.
As a result, estimating interference at the batch level rather than at the kernel level may offer a tractable solution.
Prior work~\cite{kim_interference-aware_2024,mendoza_interference-aware_2021,choi_serving_2022,yeung_horus_2022,kim_interference-aware_2021, zhao_ml_2025} has trained an ML model to estimate kernel execution interference.
However, several limitations remain.
First, they are often restricted to pairwise or rigid co-location, neglecting scenarios involving multiple concurrent batches.
Moreover, they are coarse-grained and disregard temporal effects and \emph{co-location dynamics}. To estimate interference, prediction models typically use the resource throughput of co-located batches as input features. However, since a batch may co-locate with varying batches during execution, identifying peers solely at the time of prediction overlooks critical past context. 
For instance, a batch might have been co-located with a previously completed task for an extended period, while only briefly overlapping with the batch present during prediction.
Given that batches exhibit varying resource demands, such prediction models may rely on inputs that are insufficiently precise to capture the actual resource pressure induced by co-located batches.
Furthermore, they are typically static, which may lead to degraded performance under changing workload characteristics or GPU types~\cite{concept_drift_evidently}.
To examine this potential degradation, we conduct the following experiment.

\begin{figure}[t]
  \centering
  \includegraphics[width=\linewidth]{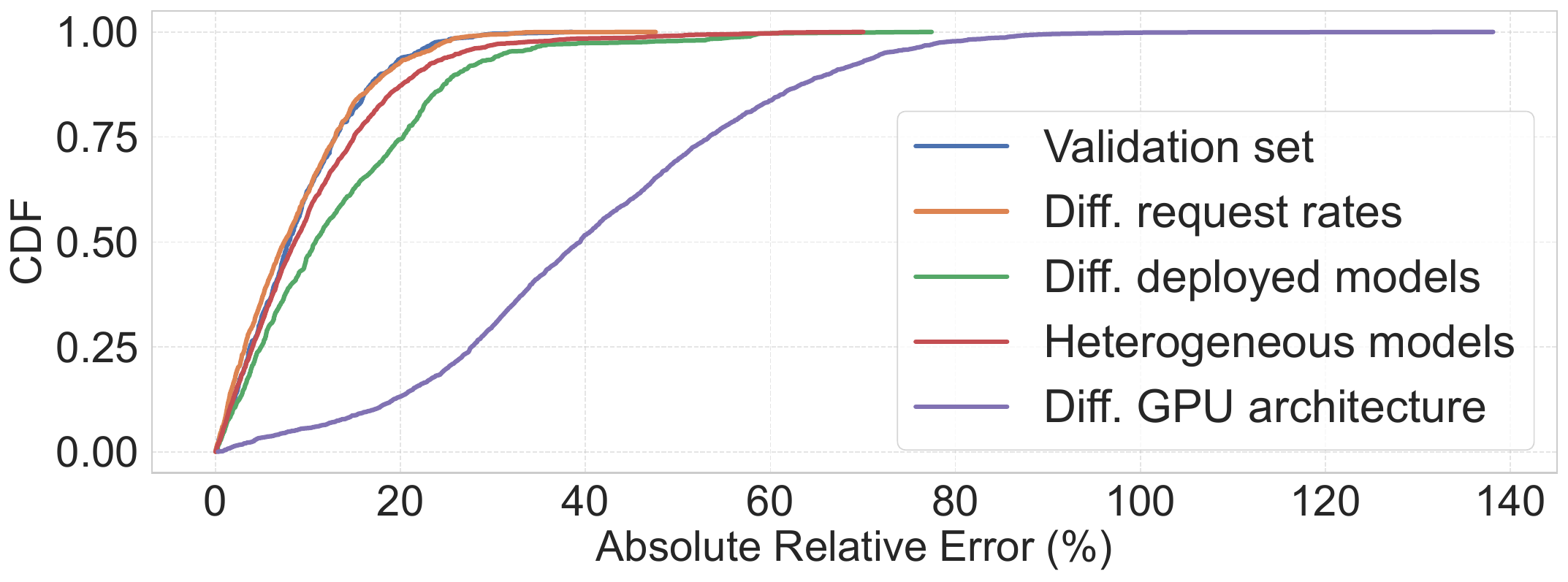}
  \caption{CDF of the prediction error under varying workload characteristics or on another GPU architecture.}
\label{figure:Strait_motivation_concept_drift}
\end{figure}

We evaluate with the \emph{gpulets}~\cite{choi_serving_2022} approach, which employs a linear regression model to estimate interference based on the resource throughputs of both the interfered batch and the co-located batch. 
The selected metrics include profiled L2 and DRAM throughputs; compute activity is ignored as \emph{gpulets} employs Multi-Process Service (MPS)~\cite{nvidia_mps} to partition compute resources. 
Therefore, we additionally include the throughput of Streaming Multiprocessors (SMs), the basic compute units, to approximate the overall compute activity.
We first deploy \emph{model set 1} (ResNet-50~\cite{he_deep_2016}, ViT-B-16~\cite{dosovitskiy_image_2021}, ConvNeXt-B~\cite{liu_convnet_2022}) and collect serving data, using 80\% of it for training and 20\% for validation. Figure~\ref{figure:Strait_motivation_concept_drift} presents the cumulative distribution function (CDF) of the absolute relative prediction error across different scenarios. 
When only the request arrival rate changes, the prediction model maintains accuracy comparable to that of the validation set.
However, when the deployed models are replaced with \emph{model set 2} (VGG-19~\cite{simonyan_very_2015}, YOLO-v8n~\cite{jocher_ultralytics_2023}, RoBERTa-B~\cite{liu_roberta_2019}), the accuracy declines noticeably, and the error distribution exhibits a longer tail.
This observation suggests that model architecture has a measurable impact on resource-contention behavior.
In a heterogeneous setting where all models are deployed, the accuracy declines, though not as significantly as in the previous case.
This occurs since the prediction model has learned the characteristics of \emph{model set 1} during the training phase, yet remains unfamiliar with those of \emph{model set 2}.
The most significant accuracy drop occurs when the prediction model is directly deployed on a different GPU architecture (NVIDIA A10G, Ampere), indicating that different GPU architectures may exhibit varying sensitivities to resource throughput. 
Several factors likely contribute to this discrepancy, including microarchitectural differences and architecture-specific compiler optimizations~\cite{lee_forecasting_2025}.
To maintain a certain degree of accuracy, the prediction model may need to adapt to the given workloads and GPU types.

\section{System Design}
\label{sec:Strait_system_design}

\begin{figure}[t]
  \centering
  \includegraphics[width=\linewidth]{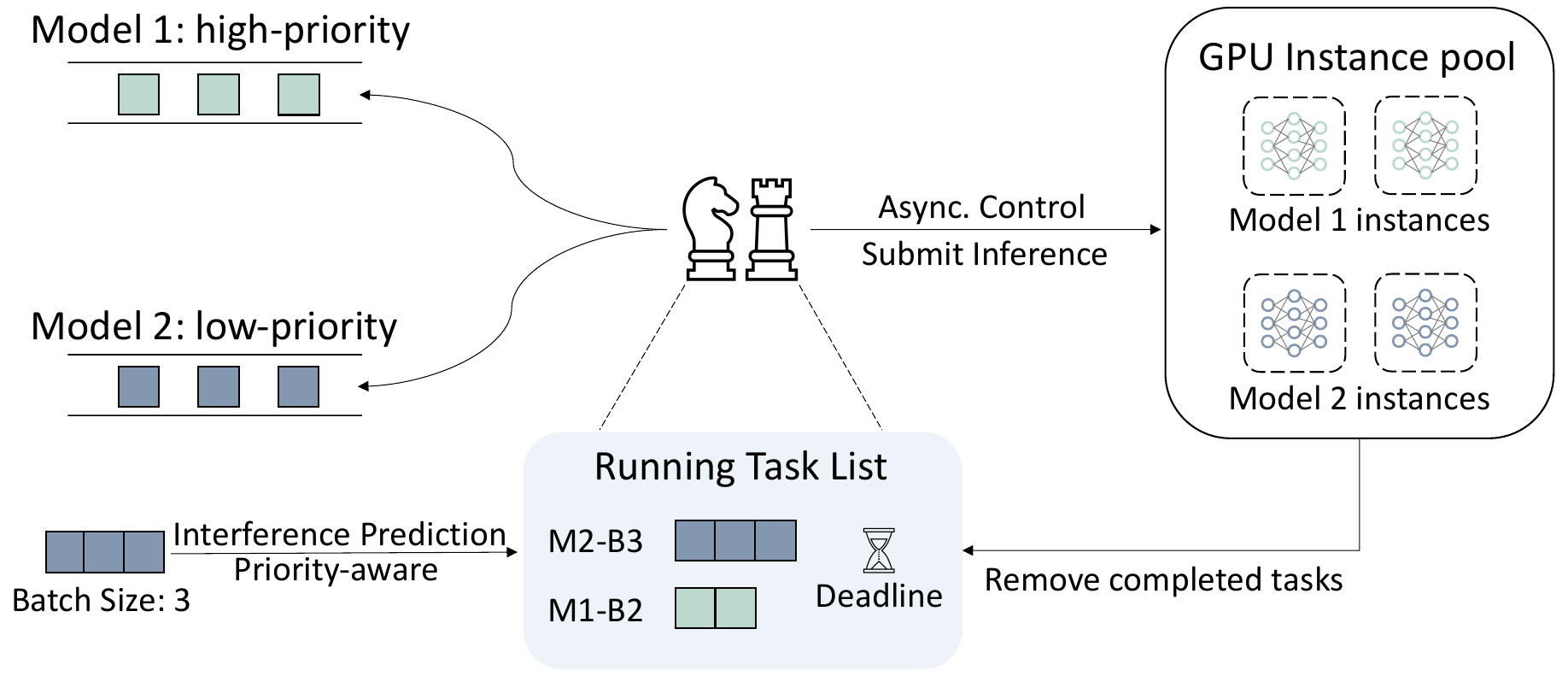}
  \caption{System design overview.}
  \label{figure:Strait_system_design_overview}
\end{figure}

We propose \emph{Strait}, an ML inference serving system that natively supports task prioritization and deadline-aware scheduling under high GPU utilization.
Similar to production inference serving systems for standard DNNs~\cite{olston_tensorflow-serving_2017,torch_serve,triton_inference_server}, Strait operates locally, enabling fine-grained control over runtime scheduling decisions to manage interference; it is thus orthogonal to cluster-level approaches and generative model solutions (Section~\ref{sec:strait_discussion}).
Figure~\ref{figure:Strait_system_design_overview} illustrates the system workflow. Incoming inference requests are initially enqueued in the front-end task queue associated with the target DNN model. 
When accessing a queue, the scheduler makes a scheduling decision and, if appropriate, dispatches the resulting batch to a backend instance pool on a designated GPU for execution.

Strait loads the DNN models and concurrently initializes their instances across all GPUs, with each model assigned a priority level (Section~\ref{subsec:Strait_priority_classification}). 
The inference latency for a batch is estimated by accounting for both interference and queuing delays. A global prediction model is used to serve all GPUs within the node to estimate kernel execution interference.
This model continuously adapts to dynamic workloads to sustain prediction accuracy (Section~\ref{subsec:Strait_interference_prediction}).
This estimated latency is then utilized in a priority-aware algorithm to guide scheduling decisions (Section~\ref{subsec:Strait_priority_aware_scheduling}).
Each GPU maintains a running task list containing GPU-specific information and details of ongoing batches for interference prediction.

\subsection{Priority Level}
\label{subsec:Strait_priority_classification}

For each model, Strait supports two priority levels: \emph{high-priority} and \emph{low-priority}. 
Strait employs CUDA stream priorities~\cite{CUDA_C++_Programming_Guide} and priority scheduling to favor HP tasks, treating LP tasks on a \emph{best-effort} basis where their performance may be sacrificed if necessary.
While this dual-priority setup aligns with previous work~\cite{strati_orion_2024,han_microsecond-scale_2022,wu_flep_2017}, it could be extended to support multiple priority levels, bounded by the GPU-specific CUDA stream priority range~\cite{CUDA_C++_Programming_Guide}.

\subsection{Interference Prediction}
\label{subsec:Strait_interference_prediction}

When a batch is submitted to a \emph{discrete} GPU for execution, it typically undergoes the following processes:
(1) Host-to-Device (upstream) data transfer: Input data is written to pinned host memory and subsequently transferred to GPU memory.
(2) Kernel execution: The GPU executes a sequence of dependent kernels and stores the resulting computations in memory.
(3) Device-to-Host (downstream) data transfer: The inference results are copied back to pinned host memory.
Accordingly, two types of resource contention may exist: one during data transfer and the other during kernel execution.

The inference latency of a batch can be estimated from its profiled isolated latency and the potential delays caused by interference and queuing.
We formalize Equation~\ref{equation:strait_overall_latency}, where $B_{i,\,j}$ represents model~\( i \) with batch size~\( j \). 
$T_{\text{inf}}(B_{i,\,j})$ is the estimated inference latency; $T_{\text{inf}}^{\text{isol}}(B_{i,\,j})$ is the profiled p95 isolated overall inference latency; $T_{\text{data}}$ accounts for the delays caused by data transfer interference; \( T_{\text{kernel}}(B_{i,\,j}) \) accounts for the delays caused by kernel execution interference; and \( T_{\text{queue}} \) denotes the queuing delay, defined as the time elapsed since the front request of the given batch was enqueued, measured at runtime.

\begin{equation}\label{equation:strait_overall_latency}
T_{\text{inf}}(B_{i,\,j}) = T_{\text{inf}}^{\text{isol}}(B_{i,\,j}) + T_{\text{data}} + T_{\text{kernel}}(B_{i,\,j}) + T_{queue}
\end{equation}

\begin{figure}[t]
  \centering
  \includegraphics[width=\linewidth]{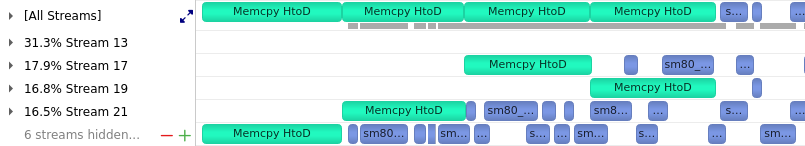}
  \caption{The figure visualizes contention during data transfer (green blocks) and kernel execution (blue blocks). When using pinned memory, concurrent batch submission results in FIFO-ordered data transfers.}
  \label{figure:Strait_data_contention}
\end{figure}

\textbf{Interference in data transfer.} This contributes to the delay term $T_{\text{data}}$.
Figure~\ref{figure:Strait_data_contention} illustrates upstream data transfer operations contending for bandwidth over PCIe, a duplex interface that handles both host-to-device and device-to-host transfers.
Given that ResNet-50~\cite{he_deep_2016} processes image inputs of dimension $3 \times 224 \times 224$, data transfer latency can constitute more than 10\% of the total inference latency, depending on the batch size, GPU type, and interconnect bandwidth.
Therefore, interference in upstream data transfer should be taken into account.
Conversely, interference in downstream data transfer may be disregarded, as many models have negligible output data sizes, and the benefits of modeling such effects may not justify the associated overhead.

Since the PCIe link handles data transfer in a FIFO manner, $T_{\text{data}}$ can be estimated based on the completion time of the previous batch.
Equation~\ref{equation:strait_data_latency} expresses the potential delays in upstream data transfer, where $t_{\text{current}}$ represents the current time and $t_{\text{available}}$ denotes the time at which the PCIe link becomes available.
Equation~\ref{equation:strait_data_available_time} defines how $t_{\text{available}}$ is updated, representing the timestamp when the PCIe link becomes available for the next batch; if $B_{i,\,j}$ is scheduled, the PCIe will be free again after the sum of its scheduled start time and its profiled p95 isolated upstream data transfer latency, $t_{\text{HtoD}}^{i, \, j}$.


\begin{equation}\label{equation:strait_data_latency}
T_{\text{data}} = \max(0,\,  t_{\text{available}} - t_{\text{current}})
\end{equation}

\begin{equation}\label{equation:strait_data_available_time}
t_{\text{available}} \leftarrow \max(t_{\text{current}},\,  t_{\text{available}}) + t_{\text{HtoD}}^{i, \, j} 
\end{equation}

\emph{Online calibration.} $T_{\text{data}}$ is estimated at scheduling time; however, it can be more precisely updated by timing the actual data transfer once it is completed.
This calibration improves the accuracy of determining whether subsequent scheduling decisions will violate the deadlines of currently executing batches (Section~\ref{subsec:Strait_priority_aware_scheduling}).
Additionally, it compensates for potential measurement skew caused by NUMA effects or differences in the PCIe Root Complex.

\textbf{Interference in kernel execution.} This contributes to the delay term $T_{\text{kernel}}(B_{i,\,j})$.
Recall that estimating interference effects at the batch level may offer a more tractable solution. 
To characterize the resource demand of a batch, we extract resource throughput metrics for its constituent kernels using Nsight Compute~\cite{nvidia_nsight_compute}. 
Assuming these kernels have resource throughputs $v_1$ to $v_n$ and corresponding execution durations $d_1$ to $d_n$, the resource demand of the batch is approximated as the time-weighted average throughput $\frac{\sum_{i=1}^{n} v_i \cdot d_i}{\sum_{i=1}^{n} d_i}$.
The delay is calculated as follows:


\begin{equation}\label{equation:strait_interference_degree}
kernel_{eff} = k \cdot b^{\mathrm{LR}
\left(
m_{\text{avg}}^{1}\,\dots\, m_{\text{avg}}^{n}
\,\mid\,
m_{\text{self}}^{\text{cmp}} \,,\, m_{\text{self}}^{\text{mem}} 
\right)} + C
\end{equation}

\begin{equation}\label{equation:strait_interference_degree_priority}
\hat{\mathit{Intf}} = 1 + kernel_{eff} \cdot \mathit{coeff}_p
\end{equation}

\begin{equation}\label{equation:strait_kernel_execution_time}
T_{\text{kernel}}(B_{i,\,j}) = \left( \hat{\mathit{Intf}} - 1 \right) \cdot T_{\text{kernel}}^{\text{isol}}(B_{i,j})
\end{equation}

Equation~\ref{equation:strait_interference_degree} estimates the interference effect, denoted as $kernel_{eff}$. 
It is parameterized by a scaling factor $k$, a base $b$, and an additive constant $C$.
The interference effect is modeled as an exponential function of \emph{increased} resource pressure, driven by the following intuition.
When a batch executes in isolation, no interference occurs. 
At lower resource pressure, interference among kernels tends to increase at a relatively slow rate. 
This is because contention for resources is effectively mitigated, as GPUs possess massive parallel resources and a throughput-oriented architecture~\cite{amert_gpu_2017, gilman_demystifying_2021, volkov_understanding_2016, zhang_sgdrc_2025}.
Furthermore, GPUs can effectively hide latency~\cite{volkov_understanding_2016} by interleaving the execution of concurrent warps (groups of threads). This mechanism allows ready warps to execute while others are stalled—e.g., during memory operations—thereby keeping the compute pipelines highly utilized.
However, as the resource pressure continues to increase, concurrency benefits diminish rapidly, since contention increasingly exceeds the GPU’s ability to mask delays, leading to rapid, super-linear slowdowns.
The exponent approximates resource pressure through a linear regression (LR) model, computed as a weighted sum of contributions from $m_{\text{avg}}^{1}$ to $m_{\text{avg}}^{n}$, $m_{\text{self}}^{\text{cmp}}$ and $m_{\text{self}}^{\text{mem}}$. 
Here, $m_{\text{avg}}^{i}$ represents the aggregate throughput of co-located batches for a specific metric used to approximate resource pressure.
Additionally, $m_{\text{self}}^{\text{cmp}}$ and $m_{\text{self}}^{\text{mem}}$ represent the compute and memory throughputs of the interfered batch itself, respectively. This reflects that batches with varying resource demands exhibit different sensitivities to resource pressure.
For $m_{\text{avg}}^{i}$, we incorporate multiple resource metrics because GPUs comprise heterogeneous execution units and a complex memory hierarchy~\cite{elvinger_understanding_2025, gpu_arch_ada_louvre}.
Specifically, to capture the overall resource pressure more accurately, we consider metrics from both the memory hierarchy (L1 cache, L2 cache, and DRAM) and the compute pipelines (tensor cores, and CUDA cores with fused multiply–add operations).
For $m_{\text{self}}^{\text{cmp}}$ and $m_{\text{self}}^{\text{mem}}$, we empirically selected metrics for the tensor cores (which specialize in dense matrix multiplications) and the L2 cache (a shared global resource for compute units) to represent these resource demands.
Recall that a batch’s set of co-located batches may change through execution; consequently, $m_{\text{avg}}^{i}$ is expected to fluctuate accordingly.
To capture this temporal dynamic—where different segments of a batch may experience varying levels of resource pressure—each time a batch arrives or departs, Strait updates $m_{\text{avg}}^{i}$ and records the corresponding timestamp.
When calculating $m_{\text{avg}}^{i}$, we employ a time-weighted average of all observed throughput values to approximate this dynamic.

Equation~\ref{equation:strait_interference_degree_priority} defines the estimated interference degree, $\hat{\mathit{Intf}}$, representing the slowdown factor, which is 1 when there is no interference.
It also captures the potential effects of CUDA stream priority through the learnable priority coefficient, $\mathit{coeff}_p$, which is uniquely assigned to each task priority level to reflect that HP tasks tend to experience less interference under identical resource pressure.
Equation~\ref{equation:strait_kernel_execution_time} estimates the \emph{delay} induced by kernel execution interference by applying the slowdown to the profiled p95 isolated kernel execution latency, $T_{\text{kernel}}^{\text{isol}}(B_{i,j})$. 

\emph{Model parameters.} The parameters can be initialized conservatively and subsequently updated via the online learning methods detailed below. Alternatively, their initial values can also be warmed up using simulated workload or restored from a checkpoint of the previous serving session.

\textbf{Self-adaptive model.}
To accommodate dynamic workloads and different GPU types (Section~\ref{sec:strait_motivation_prediction}), we dynamically update the prediction model.
Upon the completion of each batch, we compare the estimated interference degree with the \emph{actual} interference degree, which is quantified as the ratio of the measured kernel execution latency to the profiled p95 isolated value.
By continuously feeding back this discrepancy, the prediction model dynamically recalibrates itself, adjusting for over- or underestimations of interference to reflect recent resource contention behavior.

To rapidly adapt to dynamic workloads and maintain stability under concept drift~\cite{concept_drift_evidently}, this regression model requires a robust optimization strategy.
Consequently, we collected and replayed evolving workload traces (e.g., varying request arrival rates and diverse model deployments) in an online fashion to evaluate candidate methods.
We restrict our selection to first-order methods, as they are computationally efficient and exhibit greater numerical stability to noise than second-order methods in online learning.
Among candidates, we select Adam~\cite{kingma_adam_2017}, an adaptive method that automatically adjusts the effective learning rate for each parameter based on past gradients.
Compared to SGD~\cite{lecun_backpropagation_1989} and SGD with momentum~\cite{qian_momentum_1999}, adaptive optimization methods exhibit greater robustness during updates; we hypothesize that this is because global learning rates are susceptible to large variances in gradient magnitudes, leading to instability. 
While Adagrad~\cite{duchi_adaptive_2011} and AMSGrad~\cite{reddi_convergence_2019} perform well empirically, their monotonically decaying learning rates reduce adaptability over time.
Since we update the prediction model upon receiving inference feedback from each completed batch, we prefer Adam over RMSProp~\cite{rmsprop}, as its formulation facilitates smoother updates.

\textit{Hyperparameter tuning}. We empirically tune the Adam hyperparameters to determine the optimal calibration step size based on the prediction error~\cite{lecture_optimizer,optimizer_summary}.
Specifically, to enhance responsiveness to recent gradients, we adjust $\beta_1$ to 0.7 and $\beta_2$ to 0.9. To balance stable updates with rapid adaptation, we empirically set the learning rate $\alpha$ to 0.0075.
Furthermore, we employ the Huber loss function~\cite{Huber_Loss} to promote stable updates, as outliers induced by severe prediction errors could otherwise trigger disproportionately large gradients that destabilize the online learning process. 
We set this threshold to 0.50, corresponding to the p95 absolute prediction error observed when fitting the model to the captured serving traces.

\begin{algorithm}[t]
\DontPrintSemicolon
\caption{Scheduling Policy}\label{alg:strait_priority_aware_scheduling}
\KwIn{$T_{\text{inf}}()$: estimated inference latency (Section~\ref{subsec:Strait_interference_prediction})}
\KwIn{$C_{low}$: capping throughput for LP tasks}
\SetKwData{RunningTasks}{running\_tasks}

\RunningTasks $\gets [\,]$\;

\For{task queues in descending priority \label{line:strait_initialize}}{
    Removing requests with early dropping\;
    
    \For{$k \gets$ \text{Binary Search}($1$ \KwTo $j$)}{
        \For{$m \gets$ GPU($1$ \KwTo $all$)}{
            \If{$\neg violate(C_{low}) \land meet(B_{i,\,k})$}{
                $bs \gets k$ \tcp*[r]{largest batch size}
                $\text{latency}[bs,m] \gets T_{\text{inf}}(B_{i,\,k})$\;
            }
        }
    }
    
    \If{$bs$ \textup{is set}}{
        $n \gets \text{argmin}(\text{latency}[bs])$\;
        Submit $B_{i,\,bs}$ to GPU $n$\;
        Log this batch to \RunningTasks\;
    }
}

\tcp{Running in background execution threads}
\While{$true$}{
    Remove the completed batch from \RunningTasks\;
    Update the prediction model\;
}
\end{algorithm}

\subsection{Priority-aware Scheduling}
\label{subsec:Strait_priority_aware_scheduling}

Algorithm~\ref{alg:strait_priority_aware_scheduling} presents the scheduling policy. The scheduled batch should potentially meet its own deadline without violating the deadlines of ongoing tasks with equal or higher priority.

The scheduler first sorts the task queues awaiting scheduling according to the priority levels of their associated models.
We employ the early-dropping mechanism~\cite{shen_nexus_2019,hu_scrooge_2021} to discard requests whose remaining time is insufficient for completion even in isolation. This mitigates cascading deadline violations and prevents memory exhaustion under overload.
When the scheduler reaches the task queue for model \(i\), assuming \(j\) requests have been buffered.
While larger batch sizes improve GPU utilization, they tend to impose stricter time budgets (longer inference latency) and may cause more severe interference with other batches; therefore, we employ binary search to determine the maximum feasible batch size that satisfies the following constraints.

\textbullet\
\texttt{violate()} evaluates whether the schedule 1) causes the aggregate resource throughput of LP tasks on a GPU to exceed the capping value $C_{low}$, or 2) violates the deadlines of ongoing batches with equal or higher priority.
Regarding condition 1), we restrict the aggregate peak resource throughput~\cite{ncu_metrics} of LP tasks on the GPU.
This prevents them from aggressively acquiring resources, as GPUs may not provide true task-level preemption~\cite{strati_orion_2024,han_microsecond-scale_2022,shen_xsched_2025,ng_paella_2023}.
To adapt to the workload, $C_{low}$ is adjusted using an additive-increase multiplicative-decrease (AIMD) policy. 
We perform a sensitivity study in Appendix~\ref{appendix:Strait_appendix_adaptive} to select the control parameters under an evolving workload. 
To balance aggressive resource acquisition of LP tasks against suboptimal GPU utilization, $C_{low}$ ranges from $75\%$ to $100\%$, with a conservative increase rate of $0.25\%$ every 100 ms. Upon an HP task deadline violation, $C_{low}$ resets to the initial value.
Regarding condition 2), we check whether the schedule violates the deadlines of ongoing batches. Notably, an HP task can still be scheduled if it causes deadline violations only for LP tasks. For the ongoing batch, we employ a time-weighted average to approximate the co-located resource throughput (Section~\ref{subsec:Strait_interference_prediction}); under the interference effects, we then estimate the completed portion based on the elapsed time. We verify deadline feasibility by assuming that the remaining portion executes under the updated co-located throughput if the new batch is scheduled.

\textbullet\
\texttt{meet()} evaluates whether the scheduled batch can potentially meet its own deadline. 
However, this batch’s prospective average co-located resource throughput remains uncertain at scheduling time.
Consequently, we conservatively require that the scheduled batch, if assigned to a GPU, can potentially meet its deadline under half of that GPU’s runtime throughput.
Under this assumption, we record the estimated latency in a $latency$ list to select the GPU yielding the lowest latency in a multi-GPU node.

If a batch is eligible for scheduling, its details are recorded in the running task list.
Otherwise, this indicates that system resources are insufficient for the model, and scheduling is deferred. The system then proceeds to the next task queue, if one exists.
When a batch completes, its execution thread removes the corresponding record from the running task list, and the measured kernel execution latency together with its estimated value is used to update the prediction model.

\section{Implementation}
\label{sec:strait_implementation}

The prototype includes \textasciitilde 5473 lines of C++ code in the critical paths of serving.
To utilize the profiling results, online serving must run at the same GPU frequency as offline profiling.
We implemented Strait using the TensorRT runtime~\cite{tensorrt_doc}. For each model, we developed a thread-pool-based instance pool where each instance maintains independent pinned memory for data transfer and dedicated resources for inference execution.
The scheduler manages a task queue for each DNN model, and each model notifies the scheduler when preset scheduling criteria are met (e.g., batch formation timeout).
A thread-safe mechanism records information about ongoing batches to enable interference prediction and latency estimation.

\section{Evaluation}
\label{sec:strait_evaluation}

We evaluate the prototype from the following perspectives: 
(\romannum{1}) comparison against existing scheduling policies and software-defined preemption approaches; 
(\romannum{2}) inaccuracy of prediction, performance gains of an adaptive model relative to a static baseline, sustainability over time, and sensitivity to profiling drift; and (\romannum{3}) ablation study of the prioritization mechanisms and an analysis of scheduling overhead.

\subsection{Methodology}
\label{sec:strait_evaluation_workload}

\noindent
\textbf{Experimental Setup.}
We evaluate Strait primarily using two types of cloud instances: a 4-GPU node (4 NVIDIA L4-24GB GPUs, 32 vCPUs, and 192 GB DRAM) and a single-GPU node (1 NVIDIA L4-24GB GPU, 8 vCPUs, and 48 GB DRAM).
To ensure consistency, we fix the GPU frequency at 1650 MHz during both profiling and serving.
The system operates on Ubuntu 22.04, with computational acceleration enabled by CUDA 12 and TensorRT 10.

\begin{table}[t]
\centering

\caption{List of DNN models used in the evaluation.}
\label{tab:strait_tab_model}
\resizebox{\columnwidth}{!}{ 
\begin{threeparttable}
  \begin{tabular}{>{\raggedright\arraybackslash}p{2cm}  c c c c c c}
  \toprule
  \textbf{Model} & \textbf{Architecture} & \textbf{Task Classification} &  \textbf{GFLOPs} &\textbf{Priority} &\textbf{Deadline} \\
  \midrule
   ResNet-50 & CNN & Image Classification & 4.09 & high & 8 ms \\
   ViT-B-16 & Transformer & Image Classification & 16.85 & high & 15 ms \\
   ConvNeXt-B& CNN & Image Classification & 15.35&  low & 25 ms\\
   VGG-19 & CNN & Image Classification & 19.63 &  low & 25 ms\\
   YOLO-v8n & CNN & Object Detection & 9.57 & low & 20 ms\\
   RoBERTa-B & Transformer & Machine Comprehension & 10.87 & low & 45 ms \\
  \bottomrule
  \end{tabular}

%

\end{threeparttable}
} 
\end{table}

\noindent
\textbf{Workloads.} 
We consider representative pre-trained DNN models that are already widely used for downstream tasks.
As listed in Table~\ref{tab:strait_tab_model}, ResNet-50~\cite{he_deep_2016}, VGG-19~\cite{simonyan_very_2015}, ConvNeXt-B~\cite{liu_convnet_2022}, and YOLO-v8n~\cite{jocher_ultralytics_2023} are classified as CNN architectures, while ViT-B-16~\cite{dosovitskiy_image_2021} and RoBERTa-B~\cite{liu_roberta_2019} are categorized as Transformer architectures.
We assign a specific priority level to each model; however, Strait \emph{permits} the deployment of the same model with a different priority by launching new instances with distinct trigger names.
We set a uniform maximum batch size of 8 per model and a batch concurrency limit of 4 per GPU.
The \emph{deadline} for each model is preset and configured based on its isolated inference latency and urgency.
Each model is assigned a \emph{batch formation timeout} proportional to its deadline; it holds the initial request briefly to await subsequent arrivals, thereby opportunistically forming a larger batch.

As in previous work~\cite{strati_orion_2024}, we evaluate Strait using \emph{stochastic} workloads modeled as a Poisson process~\cite{cao_nonstationarity_2001, reddi_mlperf_2020}, \emph{uniform} workloads~\cite{nuscene_data_collection}, and production serverless traces~\cite{shahrad_serverless_2020}.
Each workload represents a heterogeneous scenario in which multiple models, each with distinct configurations and priority levels, may be triggered at varying arrival rates.
We adopt the open model setting~\cite{schroeder_open_2006}, in which each request is generated independently of prior inference completions.

\noindent
\textbf{Baselines.}
We reimplemented the scheduling policies of other serving systems, where the \emph{scheduling unit} is at the request level.
All policies incorporate CUDA stream priorities, task priority scheduling, and an early-dropping mechanism.
\\(1) Temporal sharing: 
This policy exclusively executes batches sequentially~\cite{crankshaw_clipper_2017, crankshaw_inferline_2020, gujarati_serving_2020, shen_nexus_2019}.
\\(2) Static spatial sharing: 
This policy caps the maximum number of concurrent batches per GPU at 3~\cite{olston_tensorflow-serving_2017,triton_inference_server}.
\\(3) Reactive spatial sharing: The policy reactively adjusts system resources based on inference feedback~\cite{dhakal_gslice_2020,romero_infaas_2021}. 
Since INFaaS~\cite{romero_infaas_2021} mitigates interference via cluster-level scaling and GSLICE~\cite{dhakal_gslice_2020} leverages MPS~\cite{nvidia_mps}—a different GPU sharing mechanism—to reallocate resources among co-located models, a direct comparison may be infeasible.
Drawing inspiration from these approaches, we dynamically limit the batch concurrency allowance for LP tasks when a model experiences severe interference, thereby mitigating contention and facilitating task prioritization. We set the global batch concurrency limit per GPU to 4 and assign a default upper bound of 3 to each task priority. If an HP task misses its deadline, the concurrency allowance for LP tasks is reduced by 1. To prevent LP tasks from being completely halted, we enforce a strict lower bound of 1. Every 200 ms, this allowance is unconditionally reset to the default upper bound of 3, providing LP tasks an opportunity to reclaim throughput.

Furthermore, we compare with software-defined preemption approaches~\cite{strati_orion_2024, shen_xsched_2025}, which intercept kernel sequences issued by applications to prevent all kernels from being submitted to the GPU at once. Consequently, their \emph{scheduling unit} operates at the kernel level.
We report this comparison in Section~\ref{subsec:Strait_evaluation_sdp}.
\\(4) XSched~\cite{shen_xsched_2025} implements abstract queues for kernels from diverse applications and enables varying degrees of preemption tailored to vendor-specific hardware capabilities. For inference serving, it employs fixed-priority scheduling~\cite{liu_scheduling_1973} on recent NVIDIA GPUs.

\noindent
\textbf{Evaluation Metrics.}
The latency of a request is measured from the moment it is enqueued in the task queue to the point when its inference result is written back to the host memory.
Requests omitted by the early dropping mechanism are also counted as deadline-violation requests.
We use \emph{goodput} to denote the number of requests meeting deadlines in a time window.

\begin{figure*}[t]
\centering
\begin{subfigure}[t]{0.32\linewidth}
\centering
\includegraphics[width=\linewidth]{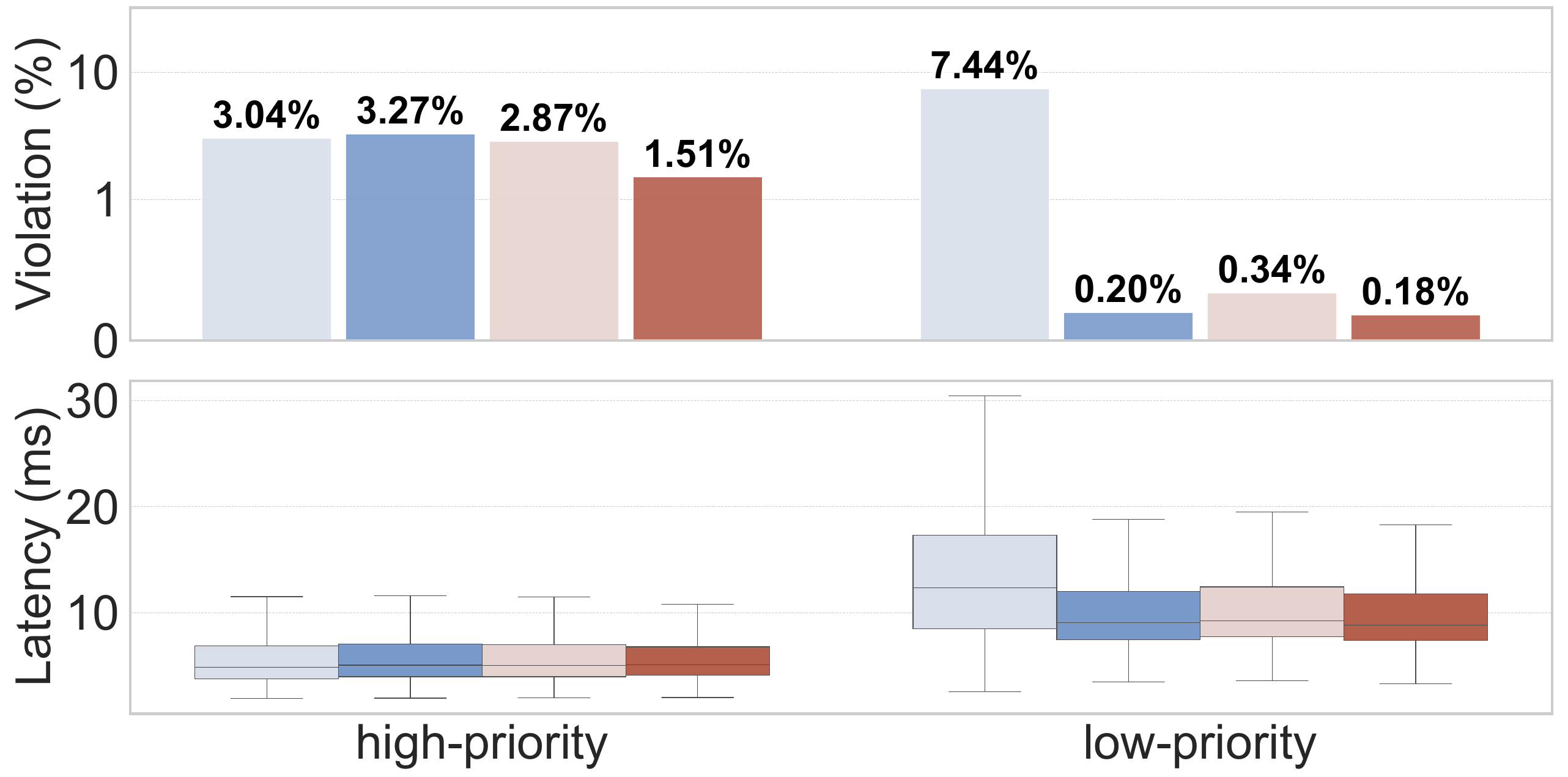}
\caption{Frequent triggers on HP tasks}
\label{figure:Strait_performance_quad_1}
\end{subfigure}
\hfill
\begin{subfigure}[t]{0.32\linewidth}
\centering
\includegraphics[width=\linewidth]{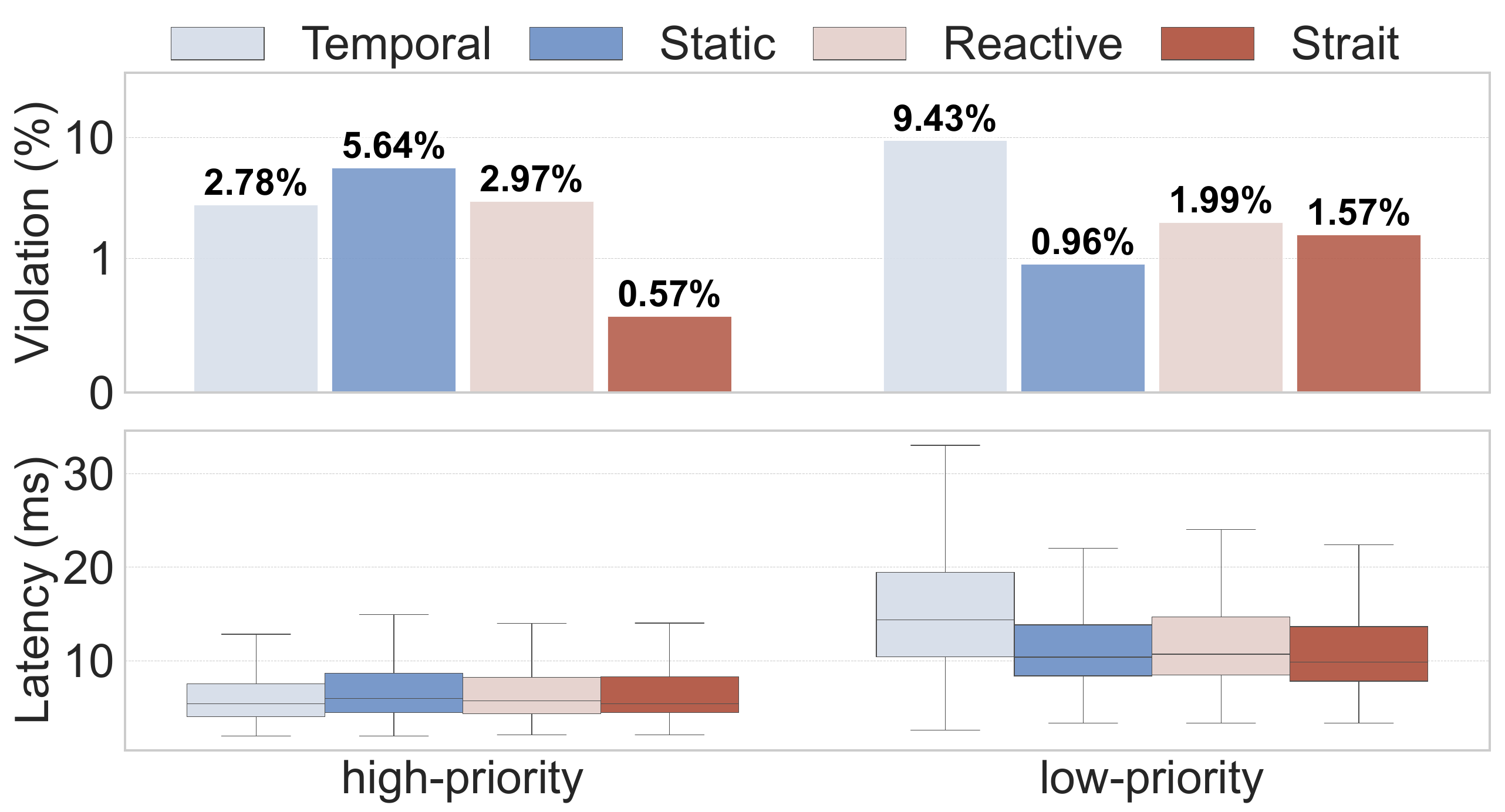}
\caption{Frequent triggers on LP tasks}
\label{figure:Strait_performance_quad_2}
\end{subfigure}
\hfill
\begin{subfigure}[t]{0.32\linewidth}
\centering
\includegraphics[width=\linewidth]{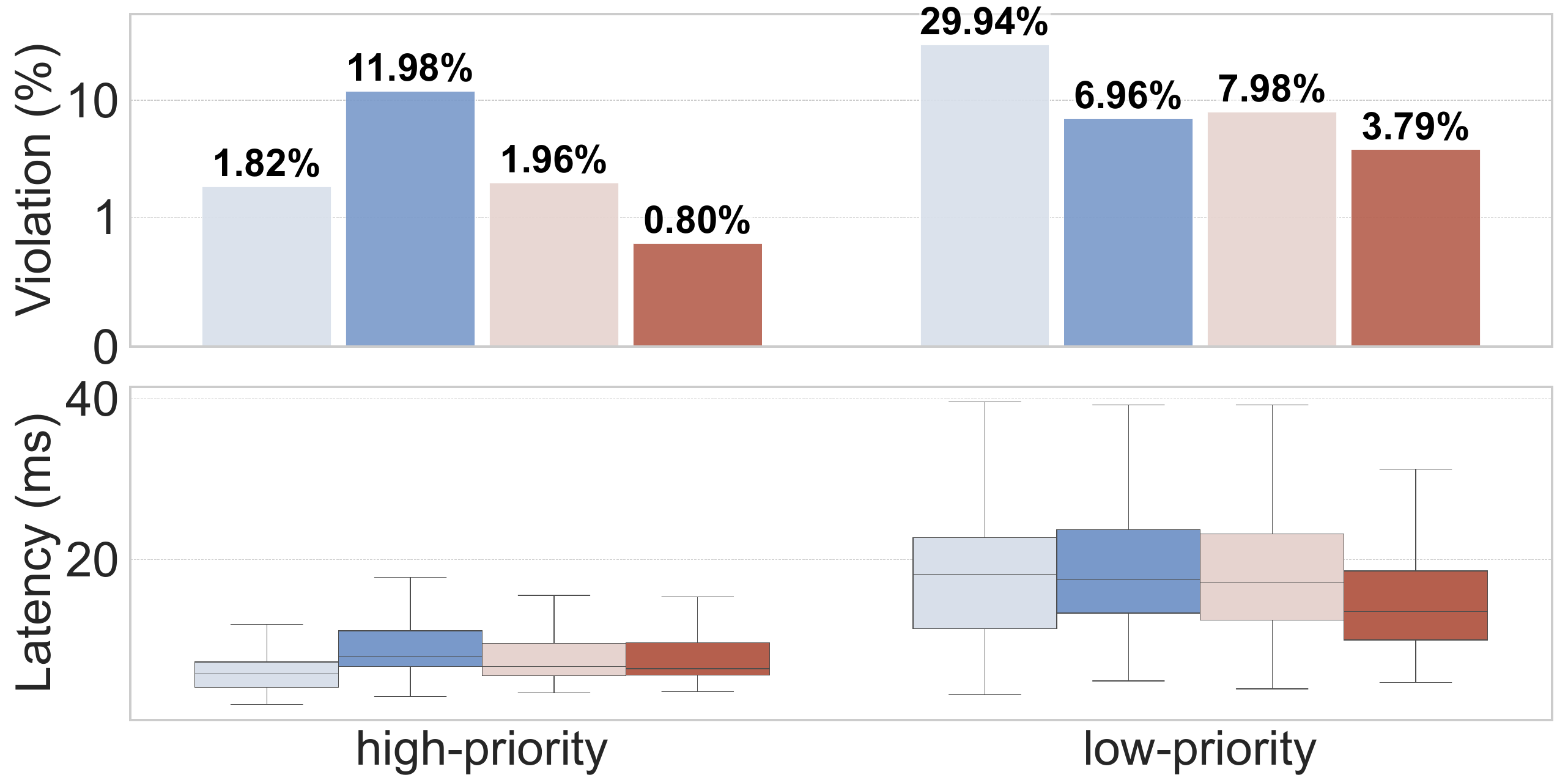}
\caption{Balanced distribution (uniform)}
\label{figure:Strait_performance_quad_3}
\end{subfigure}
\caption{Deadline violation rates and latency distributions for different scheduling policies under a 4-GPU node.}
\label{figure:strait_evaluation_performance}
\end{figure*}

\subsection{Comparison with Request-Level Scheduling}
\label{subsec:Strait_evaluation_handling_heterogeneous_workloads}

We synthesize both stochastic and uniform workloads to evaluate these policies. Motivated by the skewed request arrival distributions observed in production traces~\cite{shahrad_serverless_2020}, we configure the stochastic workloads to frequently trigger either HP or LP tasks. Consistent with uniform workloads used in monitoring scenarios~\cite{nuscene_data_collection}, we balance the requests across models. To emulate more heterogeneous scenarios, we duplicate each evaluated model listed in Table~\ref{tab:strait_tab_model}, yielding a total of 12 tasks. To evaluate performance under high GPU utilization, we use intense workloads and \emph{intentionally} induce deadline violations for quantitative comparison. Figure~\ref{figure:strait_evaluation_performance} illustrates the comparative performance.

Temporal sharing exhibits a high deadline violation rate.
It suffers from severe HOL blocking, especially for LP tasks; most requests are dropped early rather than incurring wasted computation (executions that fail to meet deadlines), resulting in low GPU utilization.
Nevertheless, it avoids interference and can immediately serve HP tasks once the GPU becomes available, as evidenced by the lower and tighter HP latency box plots (e.g., Figure~\ref{figure:Strait_performance_quad_3}). This allows temporal sharing to outperform spatial sharing baselines in terms of task prioritization in certain scenarios.

Static spatial sharing improves GPU utilization but exhibits poor prioritization, primarily due to wasted computation.
Because it completely ignores interference and co-location dynamics during scheduling, its performance is highly sensitive to workload characteristics.
For instance, it incurs a significant increase in deadline violations for HP tasks under uniform workloads.
Nevertheless, it may achieve the best performance for LP tasks by mitigating severe HOL blocking and employing fewer prioritization techniques than both the reactive approach and Strait.

While reactive spatial sharing improves task prioritization over the static approach, its effectiveness remains highly sensitive to workload characteristics. 
When HP requests dominate, the improvement for HP tasks is marginal (0.4 pp), whereas other scenarios yield gains of up to 10.02 pp. 
This discrepancy arises because the reactive approach facilitates prioritization primarily by throttling LP tasks. 
Consequently, its effectiveness is limited when LP requests are infrequent (Figure~\ref{figure:Strait_performance_quad_1}). 
Moreover, since it still lacks explicit interference management, it remains unable to effectively mitigate wasted computation. 
Notably, this prioritization may overly penalize LP tasks by blocking them, resulting in the lowest deadline satisfaction among spatial sharing approaches.

Across the evaluated scenarios, Strait reduces the deadline violations for HP tasks by 1.02 pp to 11.18 pp; however, it underperforms compared to the strongest baseline for LP tasks by 0.61 pp in the worst case. Several factors contribute to these outcomes. First, Strait employs interference prediction to facilitate deadline-aware scheduling; as evidenced by a tighter latency distribution compared to other spatial-sharing approaches, this effectively reduces wasted computation.
Nevertheless, its scheduling policies may compromise LP task performance. 
Specifically, Strait schedules HP tasks even if they cause ongoing LP tasks to violate their deadlines. Conversely, LP tasks are blocked if their execution risks violating the deadlines of ongoing HP tasks. Additionally, it throttles LP tasks using an adaptive cap.
An analysis of the causes of LP deadline violations reveals that approximately two-thirds of the violating requests result from early dropping rather than wasted computation.
Since LP tasks are treated as \emph{best-effort}, the observed degradation of less than 0.61 pp may represent an acceptable performance trade-off.

\subsection{Comparison with Kernel-Level Scheduling}
\label{subsec:Strait_evaluation_sdp}

We use XSched~\cite{shen_xsched_2025} to support fixed-priority scheduling~\cite{liu_scheduling_1973} for Triton inference server~\cite{triton_inference_server}.
We adopt XSched's original configuration, where batching is not employed and kernels are directly submitted to its abstract queues. For comparison, we enable batching in Strait but eliminate the batch formation timeout. Both systems utilize the TensorRT~\cite{tensorrt_doc} runtime.
To account for Triton-introduced overhead (e.g., gRPC), we measure the difference in isolated inference latency between Triton and Strait for each model. We then subtract this difference from the measured latency of XSched to ensure a fair comparison.

\begin{figure}[t]
  \centering
  \begin{subfigure}[t]{0.49\linewidth}
    \centering
    \includegraphics[width=\linewidth]{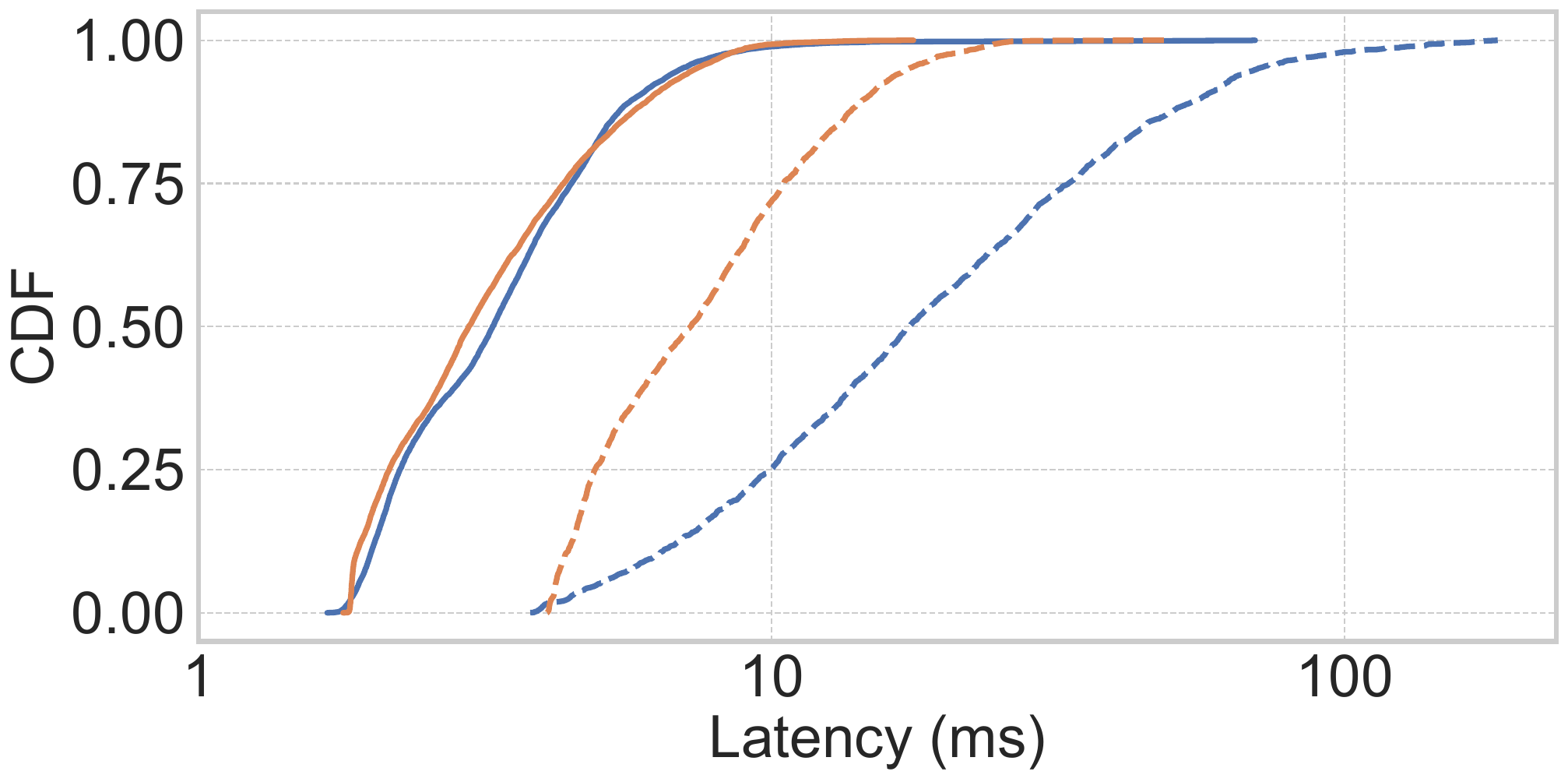}
    \caption{Pressure on HP tasks}
    \label{figure:Strait_evaluation_software_preemption_a}
  \end{subfigure}
  \hfill
  \begin{subfigure}[t]{0.49\linewidth}
    \centering
    \includegraphics[width=\linewidth]{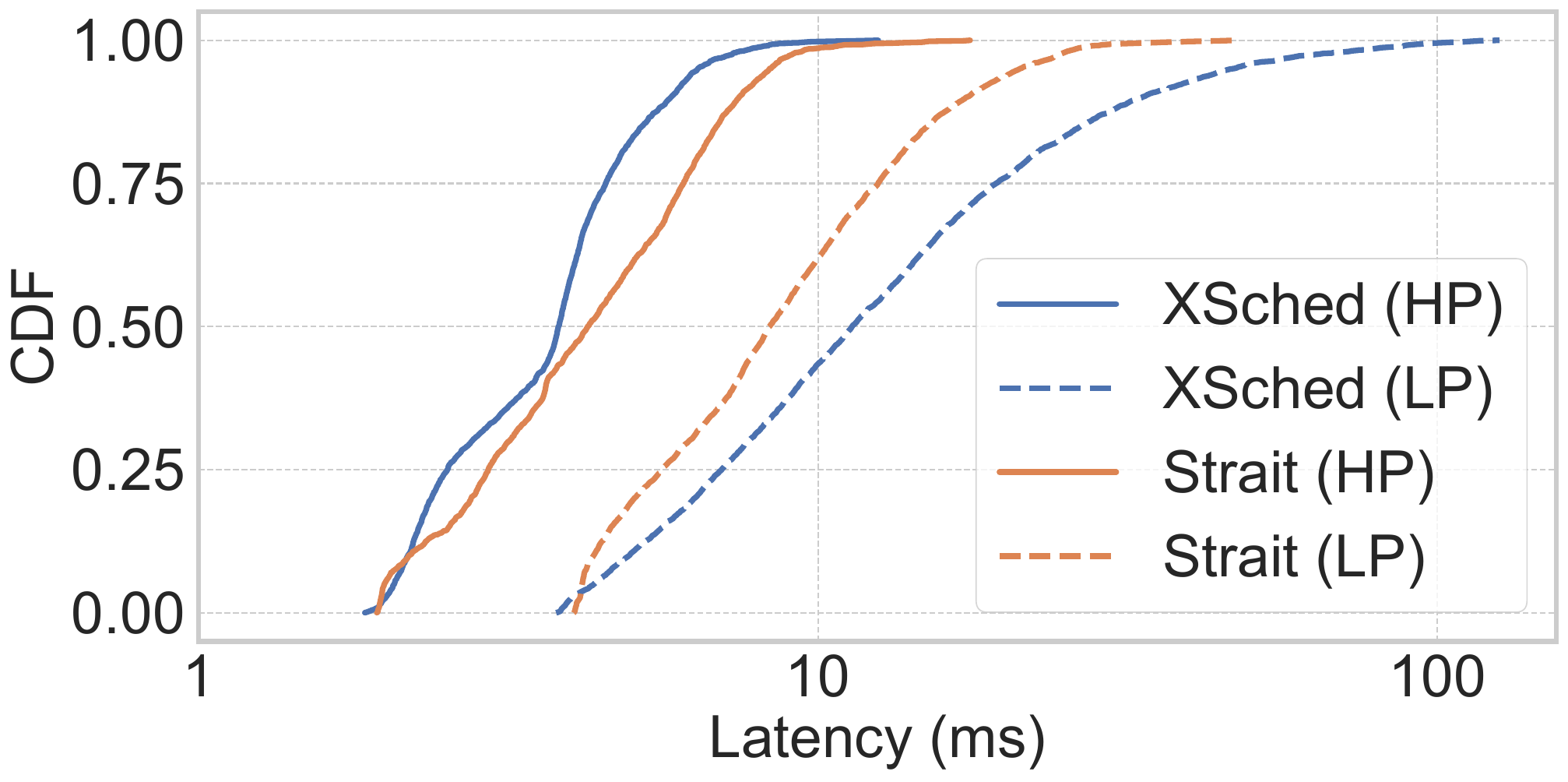}
    \caption{Pressure on LP tasks}
    \label{figure:Strait_evaluation_software_preemption_b}
  \end{subfigure}
  \caption{CDF of inference latency for XSched and Strait.}
  \label{figure:Strait_evaluation_software_preemption}
\end{figure}

Figure~\ref{figure:Strait_evaluation_software_preemption} illustrates the inference latency distributions. 
When pressure is on HP tasks (case in Figure~\ref{figure:Strait_evaluation_software_preemption_a}), Strait incurs $1.43\%$ and $1.20\%$ deadline violations for HP and LP tasks, respectively, compared to $2.07\%$ and $29.89\%$ for XSched.
When pressure is on LP tasks (case in Figure~\ref{figure:Strait_evaluation_software_preemption_b}), Strait incurs $0.69\%$ and $4.27\%$ violations for HP and LP tasks, respectively, whereas XSched results in $0.043\%$ and $11.41\%$.
Overall, Strait maintains comparable deadline satisfaction for HP tasks while yielding measurable improvements for LP tasks. 
Several factors contribute to these results. 
First, XSched's prioritization allows for fine-grained control by stalling kernels from LP task launches when HP kernels arrive. 
While this mechanism effectively protects HP tasks when LP requests dominate, its efficacy degrades when HP requests dominate.
This is because equal-priority kernels can still contend for shared GPU resources, and XSched lacks explicit interference management.
Additionally, XSched may suffer from HOL blocking for LP tasks, as HP kernels delay LP kernel launches, suppressing concurrency and underutilizing GPU resources. 
In contrast, Strait can make more informed decisions based on task deadlines, such as opportunistically scheduling large batches or dropping requests that would violate deadlines, thereby achieving more equitable performance.

\begin{figure*}[t]
  \centering
  \begin{subfigure}[b]{0.32\linewidth}
    \centering
    \includegraphics[width=\linewidth]{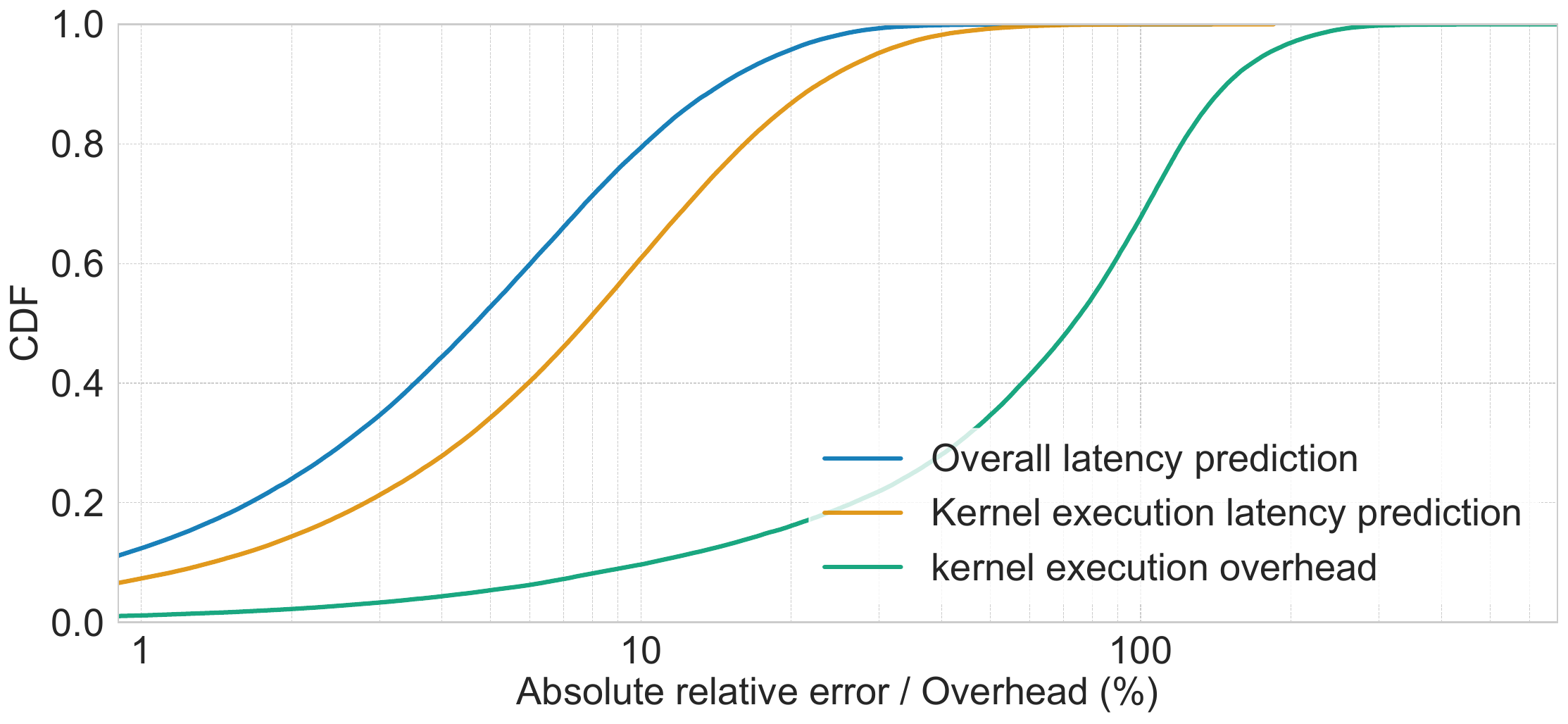}
    \caption{CDF of prediction errors and kernel execution overhead under intense loads.}
    \label{figure:Strait_evaluation_inaccuracy}
  \end{subfigure}\hfill
  \begin{subfigure}[b]{0.32\linewidth}
    \centering
    \includegraphics[width=\linewidth]{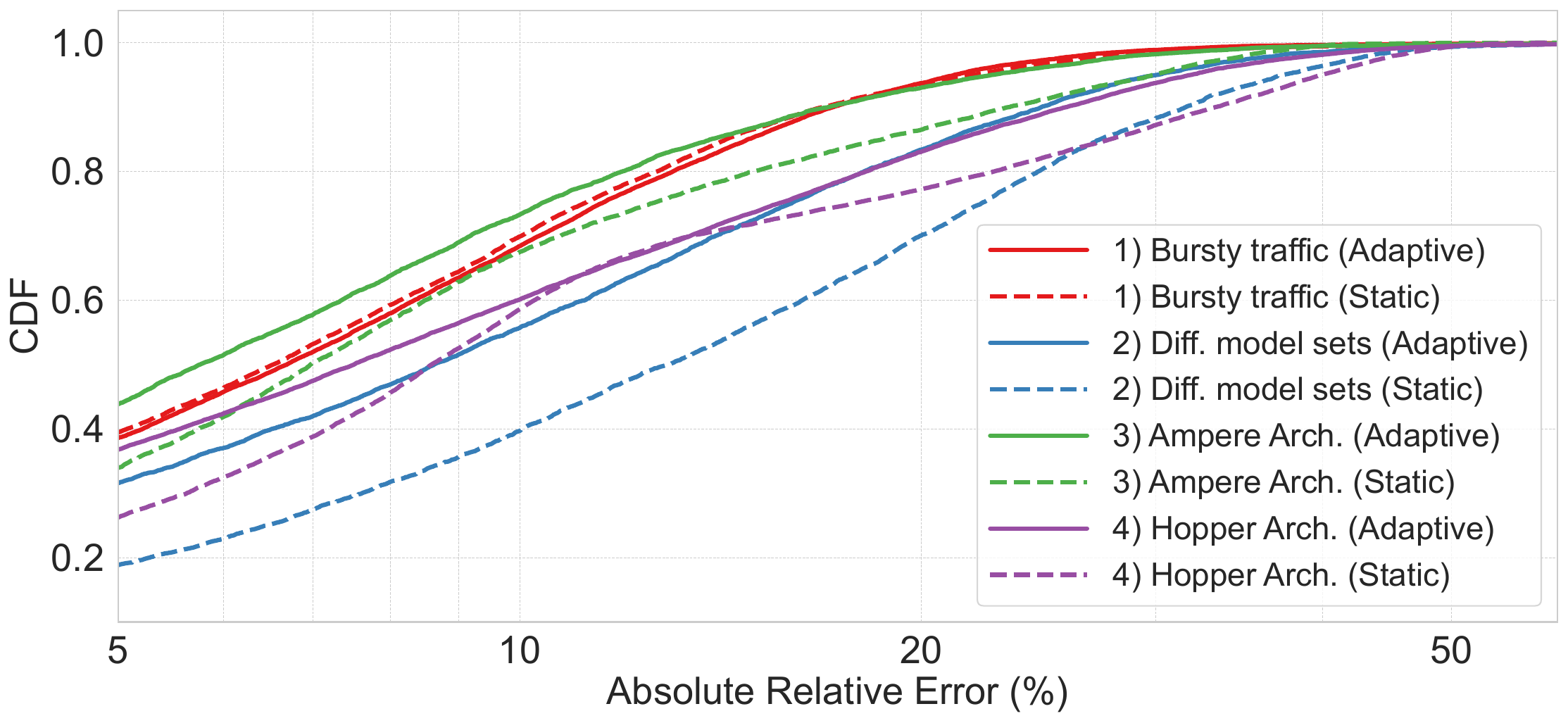}
    \caption{CDF of prediction errors with adaptive and static model under different scenarios.}
    \label{figure:Strait_evaluation_adaptability}
  \end{subfigure}\hfill
  \begin{subfigure}[b]{0.32\linewidth}
    \centering
    \includegraphics[width=\linewidth]{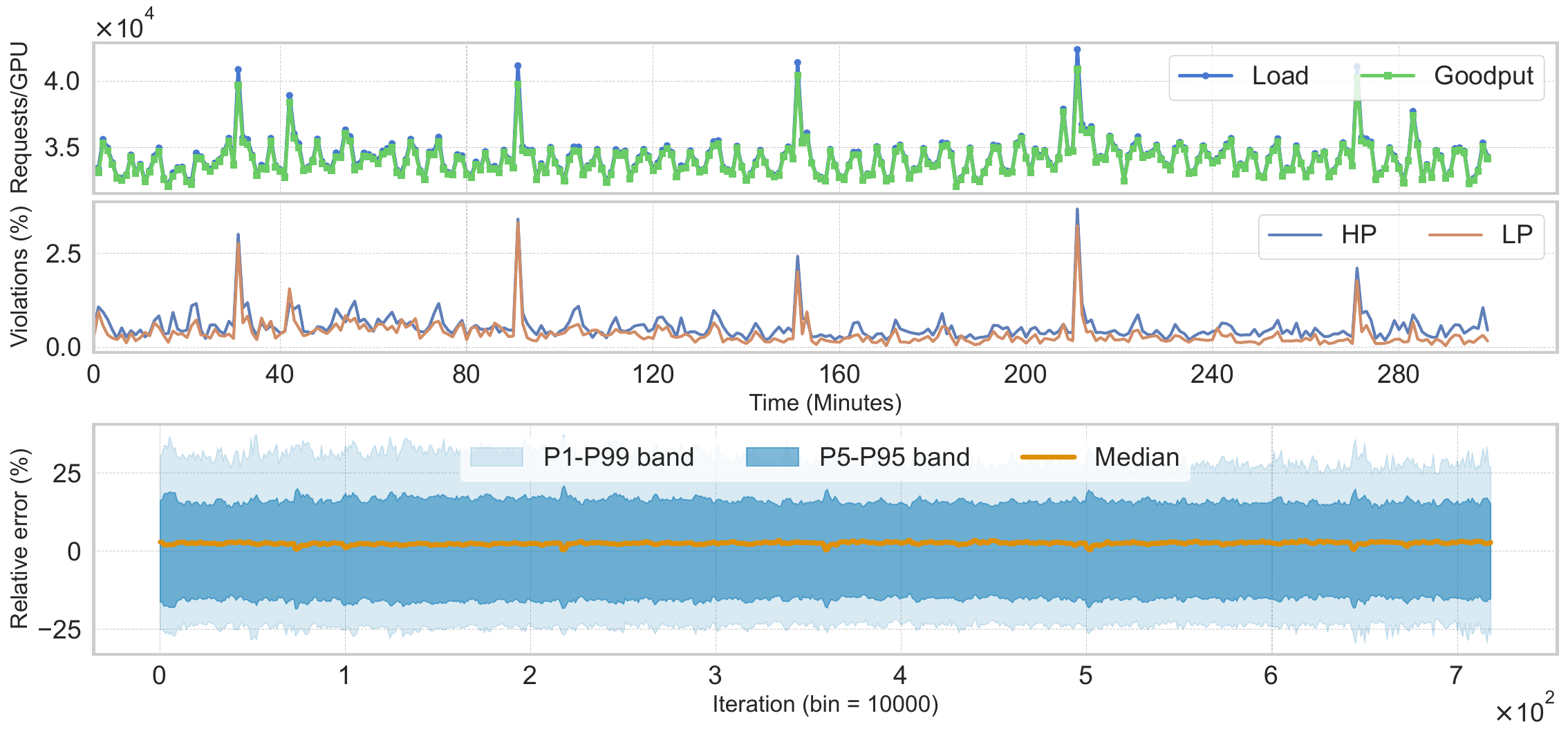}
    \caption{Serving under realistic production traces across 5 hours.}
    \label{figure:Strait_stability}
  \end{subfigure}
  \caption{Evaluation results for inaccuracy, adaptability, and sustainability.}
  \label{figure:Strait_evaluation_trace}
\end{figure*}

\subsection{Validation of Interference Prediction}
\label{subsec:Strait_validation}


Recall that we employ an adaptive prediction model to continuously recalibrate for the over- and under-estimation of interference.
We quantify interference-induced overhead in kernel execution as $\frac{\lvert {T_{\text{kernel}}^{\text{actl}}(B_{i,j})} - {T_{\text{kernel}}^{\text{isol}}(B_{i,j})} \rvert}{{T_{\text{kernel}}^{\text{isol}}(B_{i,j})}}$, where ${T_{\text{kernel}}^{\text{actl}}(B_{i,j})}$ denotes the measured kernel execution latency, and ${T_{\text{kernel}}^{\text{isol}}(B_{i,j})}$ denotes the corresponding profiled p95 value.

\textbf{Inaccuracy under intense workloads.}
Figure~\ref{figure:Strait_evaluation_inaccuracy} presents the results for the workloads used in Section~\ref{subsec:Strait_evaluation_handling_heterogeneous_workloads}, illustrating the prediction errors for both kernel execution and overall inference latency, alongside the distribution of kernel execution overhead.
The distribution of kernel execution overhead exhibits a long tail—reaching approximately 682\% with a p99 value of 245\%—which indicates that severe interference occurs during heterogeneous model deployment under GPU oversubscription.
This skewness in kernel execution overhead complicates latency prediction.

The predictions for kernel execution latency exhibit absolute relative errors of 29.6\% and 46.6\% at p95 and p99, respectively. 
Since the prediction model relies on continuous inference feedback for online updating, it is constrained by the underlying data distribution.
Due to data sparsity, the prediction model is ill-equipped to handle severe long-tail interference, primarily resulting in interference underestimation.
Moreover, this severe interference would encourage the model to calibrate for these outliers, leading to overestimation in other scenarios; e.g., by analyzing data, we find that predictions under relatively small kernel execution overhead (<60\%) tend to overestimate interference.
In contrast, we find that the model performs relatively well at an overhead of near 80\%; this is because sufficient training data is available at that range, as reflected by the steep ascent of the green curve. 
We discuss potential ways to mitigate the long tails in Section~\ref{sec:strait_discussion}.
Other factors may also limit prediction accuracy, including workload characteristics that the adaptive model does not adequately capture initially.
Additionally, we employ time-weighted averages to approximate resource pressure; however, this approach may be limited if a batch's resource demands vary significantly during execution.
Fortunately, the overall latency prediction can mitigate these prediction errors, and this value is ultimately used for scheduling.
This is because kernel execution latency is only one element of the overall latency, and we employ methods such as offline profiling and online calibration to enhance the estimation of the other elements.
As a result, the overall latency prediction errors are reduced to 18.95\% at p95 and 27.9\% at p99 in the \emph{extreme} case.

\textbf{Adaptive over static model.}
We evaluate whether an adaptive prediction model outperforms a static model under dynamic workloads or when deployed directly on other NVIDIA GPU types. 
We begin by deploying a CNN-based model set (ResNet-50, ConvNeXt-B, and YOLO-v8n) and updating the prediction model under a moderate load to initialize the model parameters. 
Subsequently, we transition to the following scenarios for 45 seconds: 1) bursty traffic; 2) an alternative model set dominated by Transformer architectures (VGG-19, ViT-B-16, and RoBERTa-B); 3) an NVIDIA A10G GPU featuring the Ampere architecture; and 4) a high-performance NVIDIA H100 GPU featuring the Hopper architecture, where loads are scaled up to introduce interference.

Figure~\ref{figure:Strait_evaluation_adaptability} presents the results. 
For scenario 1), the adaptive method demonstrates comparable, albeit slightly inferior, performance to the static model for small prediction errors (under 20\%). 
This occurs because the data distribution shifts toward regimes with more severe interference. 
Consequently, the adaptive method learns this trend, achieving around a 3 pp improvement in p99 prediction error over the static approach, though it suffers from fluctuations in smaller errors. 
For scenario 2), the adaptive model exhibits noticeable improvements for both small and tail errors (e.g., a 7 pp improvement in p95 errors), suggesting that varying the model sets may alter resource contention behavior. 
For scenarios 3) and 4), the adaptive method also yields consistent improvements across both small and tail errors, achieving reductions of over 7.2 pp and 8 pp in p99 prediction errors, respectively. This indicates that the throughput-interference mapping may vary across GPU architectures.
Accordingly, the performance gains of adaptive methods are scenario-dependent, with more noticeable improvements observed when transitioning across different model sets or GPU architectures.

\textbf{Serving a period of time.} 
Following prior work~\cite{strati_orion_2024,gujarati_serving_2020}, we evaluate Strait on production serverless traces~\cite{shahrad_serverless_2020} to assess its ability to maintain performance and prediction accuracy.
Since the traces record per-minute aggregate requests for each function, we generate discrete arrivals using a Poisson process parameterized by these rates.
To pressure Strait, we map the arrival rates of the six most frequent functions during the busiest 5-hour window of a single day to the evaluated models, scaling them according to GPU capacity.
Figure~\ref{figure:Strait_evaluation_trace} presents the results. 
Goodput generally overlaps with the incoming load, and Strait can absorb moderate load spikes (e.g., shortly after 40 mins and 280 mins), as evidenced by the minor fluctuations in deadline violation rates. 
During peak burst periods (e.g., around 90 mins), the load may exceed system capacity, leading to temporary spikes in deadline violations; however, Strait can recover shortly thereafter.
To evaluate prediction accuracy, we analyze a 10,000-batch update window to align with the throughput per GPU. The median error remains consistently near 0\%, and the two error bands exhibit steady variance. This suggests that Strait continuously calibrates for both overestimations and underestimations of interference. Although severe interference can occur when GPU resources are oversubscribed, mechanisms such as the Adam optimizer and Huber loss can prevent catastrophic updates from destabilizing the system.

\textbf{Profiling drift.} 
Since inference latency can be rapidly reprofiled whereas resource throughput profiling is significantly more time-consuming, we evaluate the model's \emph{sensitivity} to throughput profiling drift. We apply random perturbations of $\pm x$\% to each profiled value. When this perturbation magnitude is below 15\%, the p99 prediction drift remains below 1.5 pp, suggesting that the adaptive model can tolerate moderate profiling drift in throughput~(Appendix~\ref{appendix:Strait_appendix_profiling}).

\textbf{Ablation Study.}
Appendix~\ref{subsec:Strait_evaluation_task_prioritizaiton} investigates the contributions of each task prioritization mechanism, suggesting that throttling LP tasks and checking whether the scheduled batch would violate the deadlines of ongoing tasks are the most critical components.

\subsection{Overheads}
\label{subsec:Strait_evaluation_overheads}

Algorithm~\ref{alg:strait_priority_aware_scheduling} employs binary search to determine the feasible batch size for a model.
Given constant-time interference prediction, the search over candidate batch sizes \( j \) and GPU number $n$ has complexity $n \cdot O(\log j)$.
In our setup, scheduling overhead is typically bounded by tens of microseconds.
In practice, we can configure each serving system instance to manage a subset of GPUs within a node for efficient scheduling.
This also mitigates potential NUMA effects and lock contention.

\section{Discussion}
\label{sec:strait_discussion}

\noindent
\textbf{Interference prediction.}
Extreme interference may arise in bursty workloads, where severe resource contention and stall-induced delays can inflate kernel execution time.
To alleviate this issue, limiting GPU utilization by imposing a maximum aggregate throughput or a maximum allowed batch concurrency are natural choices.
However, our empirical results indicate that naively applying these limits could reduce deadline satisfaction.
We suggest that more fine-grained overload control mechanisms should be explored.

\noindent
\textbf{Priority and Scalability.}
Strait schedules inference requests on available non-preemptive GPUs, making its approach orthogonal to priority-handling mechanisms in CPU resource management~\cite{tirmazi_borg_2020,Hadoop_YARN,boutin_apollo_2014,Kubernetes} or priority-aware autoscaling~\cite{jeon_house_2025}.
It manages interference at the node level; future extensions may include priority-aware admission control and load balancing at the cluster level.

\noindent
\textbf{Generalize to edge devices.}
Strait is evaluated on server-grade discrete GPUs. However, edge devices also serve inference workloads, and they are typically equipped with integrated GPUs that share physical memory with the CPU~\cite{yan_polythrottle_2024, zhang_edgenn_2023, jayanth_benchmarking_2024, wang_unified_2019, orin_series, wu_machine_2019}. 
Although there is no explicit data transfer between host and device memory, memory contention could become more unpredictable owing to concurrent CPU activities. 
Additionally, the compute capability of integrated GPUs is more constrained and is further influenced by factors such as power and thermal budgets.
Moreover, as integrated GPUs incorporate more heterogeneous processing units, extending the input features of the prediction model could be further explored.

\noindent
\textbf{Large Language Models (LLMs).} 
Recent serving frameworks target autoregressive generation in LLMs, focusing on continuous batching, KV-cache management, and disaggregated prefill and decoding phases~\cite{yu_orca_2022, kwon_efficient_2023, zhong_distserve_2024,dynamo_triton}.
In contrast, Strait focuses on on-premises deployments~\cite{vaish_case_2023, jin_glass_2023, tran_human_2023, noghabi_emerging_2020, davari_predictive_2021} of standard DNNs with fixed input/output shapes.
In addition, on-premises scenarios may need to co-locate heterogeneous models on the same hardware, whereas LLM serving infrastructures focus on scaling one or several LLMs across multiple servers~\cite{zhong_distserve_2024}.

\section{Related Work}
\label{sec:strait_related_work}

Serving systems such as Clipper~\cite{crankshaw_clipper_2017}, InferLine~\cite{crankshaw_inferline_2020}, Clockwork~\cite{gujarati_serving_2020}, and Nexus~\cite{shen_nexus_2019} optimize inference through batch size control and framework support but uniformly rely on temporal sharing.
Serving systems that employ static spatial sharing, such as TFS~\cite{olston_tensorflow-serving_2017}, Triton~\cite{triton_inference_server}, and Scrooge~\cite{hu_scrooge_2021}, require preset concurrency or model instances, limiting adaptability to dynamic workloads.
Reactive spatial-sharing systems, such as INFaaS~\cite{romero_infaas_2021} and GSLICE~\cite{dhakal_gslice_2020}, passively mitigate interference by reallocating resources, switching model variants, or scaling instances only after observing performance degradation.
With proactive spatial-sharing, serving systems dynamically make scheduling decisions.
gpulets~\cite{choi_serving_2022} employs a linear regression model to estimate interference during resource provisioning; however, its interference prediction approach is coarse-grained and may be susceptible to evolving workloads (Section~\ref{sec:strait_motivation_prediction}).

GPU sharing techniques~\cite{han_microsecond-scale_2022, strati_orion_2024, zhang_sgdrc_2025,shen_xsched_2025,coppock_lithos_2025} enable fine-grained scheduling for task prioritization.
REEF~\cite{han_microsecond-scale_2022} enables preemption of best-effort kernels for real-time tasks, but it is limited to AMD GPUs.
Orion~\cite{strati_orion_2024} schedules kernels to avoid co-locating those with similar resource intensities while prioritizing HP tasks.
XSched~\cite{shen_xsched_2025} supports varying preemption levels depending on the accelerators' hardware capabilities; we compare it against Strait when integrated with the inference server.
LithOS~\cite{coppock_lithos_2025}, a recent work designed for inter-process applications, argues that scheduling granularity can be further refined to the thread-block level to facilitate task prioritization.
Orthogonal to these works, Strait employs a retrofitted interference prediction approach for inference workloads~\cite{kim_interference-aware_2024,mendoza_interference-aware_2021,choi_serving_2022,zhao_ml_2025,kim_interference-aware_2021} and integrates it into scheduling to enhance task prioritization.


\section{Conclusion}

ML inference serving systems are often required to support task prioritization and to deliver timely predictions under high GPU utilization.
Strait responds to this challenge by explicitly managing interference during its priority-aware scheduling. 
Notably, the interference prediction approach is adaptive to mitigate the potential degradation in prediction accuracy under changing workload characteristics and GPU types. 
The evaluation results demonstrate consistent performance gains in task prioritization and deliver more equitable performance compared to software-defined preemption approaches.
The interference prediction model serves as a key building block; we anticipate that further refinements to this model could further improve deadline satisfaction.


\bibliographystyle{ACM-Reference-Format}
\bibliography{bibliography/strait_misc,bibliography/Systems,bibliography/latency}

\clearpage
\appendix
\section{Appendix}
\subsection{Adaptive Throttling}
\label{appendix:Strait_appendix_adaptive}

\begin{figure}[t]
  \centering
  \includegraphics[width=\linewidth]{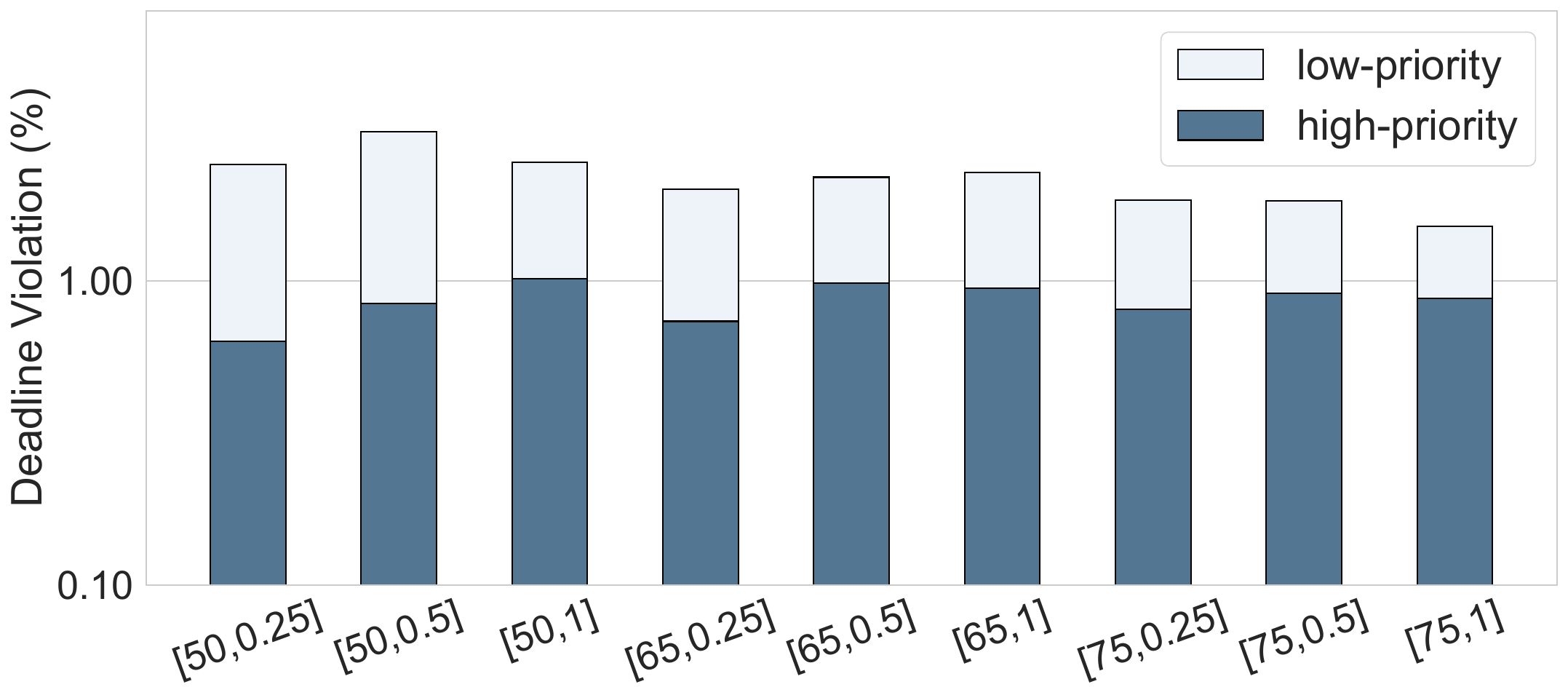}
  \caption{Deadline violation rates under various configurations. For a configuration [a, b], the first term \emph{a} denotes the initial value, that is, the lowest allowed capping value for LP tasks, whereas \emph{b} denotes the associated increase rate applied every 100 ms.}
  \label{figure:Strait_design_low_cap}
\end{figure}

We employ an AIMD policy to throttle the aggregate resource throughput of concurrent LP tasks on the GPU (Section~\ref{subsec:Strait_priority_aware_scheduling}). 
We select the control parameters to prevent LP tasks from oversubscribing GPU resources without causing severe underutilization.
Figure~\ref{figure:Strait_design_low_cap} shows the results of our sensitivity study. 
Although a lower initial value appears to improve task prioritization, it risks severely penalizing LP tasks, particularly when $a = 50$. 
Conversely, a higher increase rate may improve deadline satisfaction for LP tasks but risks reducing task prioritization. To balance these trade-offs, we adopt a larger initial value combined with a conservative increase rate. 
Specifically, $C_{low}$ ranges from 75\% to 100\%, with an increase rate of 0.25\% every 100 ms. 
When HP tasks miss their deadlines, $C_{low}$ returns to 75\%.

\subsection{Sensitivity to Profiling Drift}
\label{appendix:Strait_appendix_profiling}

\begin{figure}[t]
  \centering
  \includegraphics[width=\linewidth]{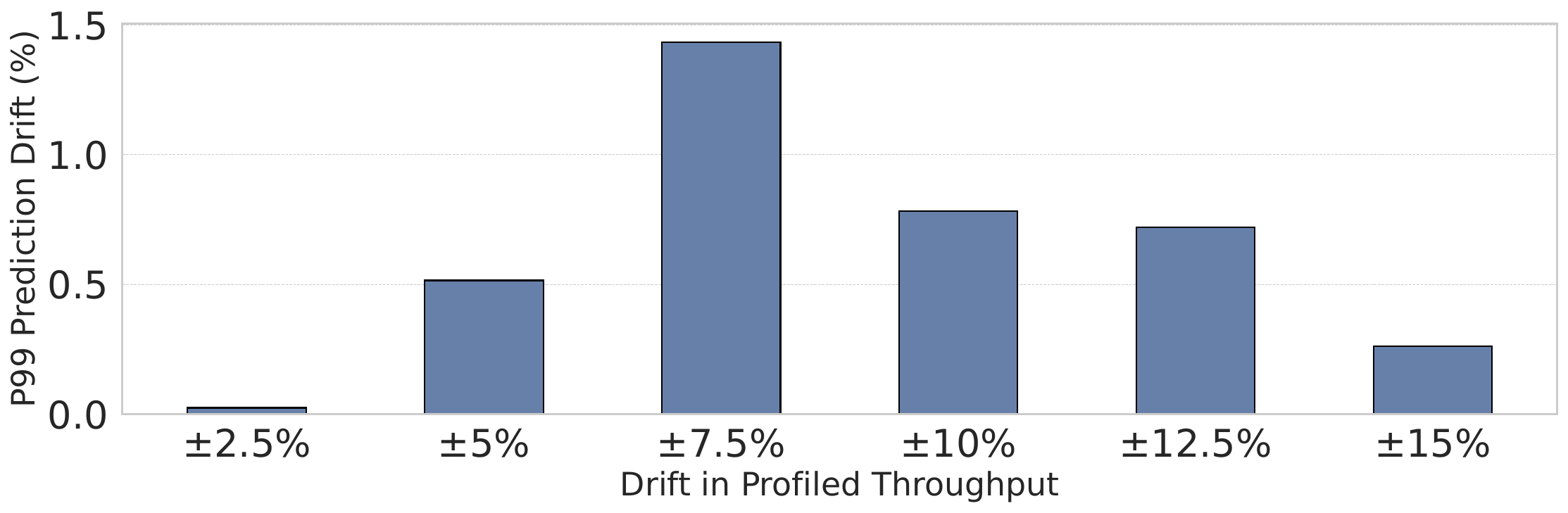}
  \caption{Profiling drift in resource throughput relative to the baseline without drift.}
  \label{figure:Strait_evaluation_profiling_drift}
\end{figure}

Strait relies on profiling data to estimate interference. We evaluate the impact when profiling data drifts due to environmental changes (e.g., different runtime versions or hardware configurations), without re-profiling the models. To simulate this, we artificially perturb the stored throughput values. We carefully control the perturbation magnitudes to ensure that all perturbed throughput values remain within valid bounds (0\%–100\%). 
We use an evolving workload to evaluate Strait's sensitivity to profiling drift. 
Figure~\ref{figure:Strait_evaluation_profiling_drift} shows that under a random perturbation rate of 15\%, the changes in p99 prediction error remain below 2 pp, suggesting that Strait can adapt to moderate levels of profiling drift. 
The small impact on prediction accuracy is attributable to Strait's online learning mechanism, which dynamically tunes the model parameters. 
Consequently, the prediction model continuously adapts to the discrepancy between the perturbed profiling data and the actual runtime behavior, thereby absorbing moderate profiling drift.
To mitigate inference latency drift, we suppose that the latency can be quickly reprofiled when necessary, or that an online monitoring mechanism can be employed to recalibrate for potential variations, provided that the batch executes in isolation.

\subsection{Ablation on Prioritization Mechanisms}
\label{subsec:Strait_evaluation_task_prioritizaiton}

\begin{figure}[t]
  \centering
  \includegraphics[width=\linewidth]{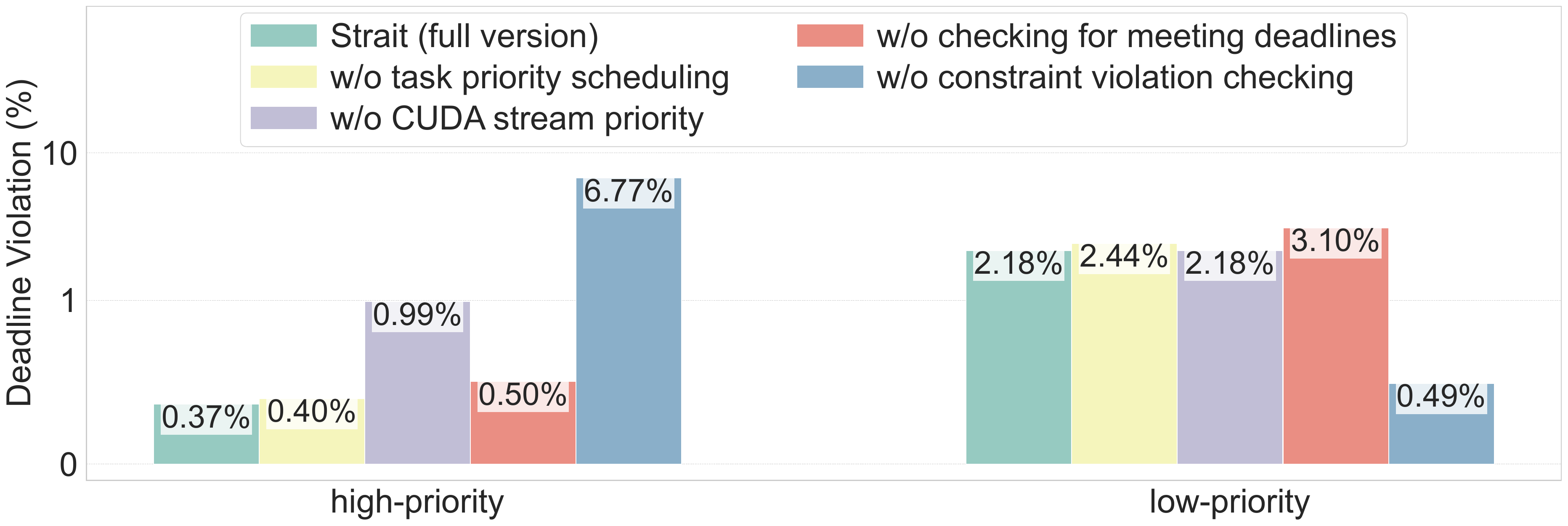}
  \caption{Impact of sequentially removing a task prioritization mechanism.}
  \label{figure:Strait_ablation}
\end{figure}

Figure~\ref{figure:Strait_ablation} presents the ablation study by gradually removing a task prioritization mechanism, evaluated under an evolving workload.
Removing task priority scheduling yields no noticeable degradation, as it neither manages interference nor enables deadline-aware scheduling.
Removing CUDA stream priority multiplies deadline violations by $2.68\times$ for HP tasks, demonstrating that hardware-level mechanisms are still useful.
Removing deadline feasibility checking multiplies deadline violations by 1.35$\times$ and 1.42$\times$ for HP and LP tasks, respectively, suggesting that leaving some headroom for potential interference and carefully selecting batch size helps improve deadline satisfaction.
The most significant impact arises from constraint violation checking.
Removing checking of deadline violations of ongoing tasks and adaptive throttling leads to an 18.3$\times$ degradation in HP tasks and a substantial improvement for LP tasks, suggesting this is the most critical component in Strait.

\end{document}